\documentclass[10pt, a4paper]{article}
\usepackage{lrec}
\usepackage{graphicx}
\usepackage{tabularx}
\usepackage{multirow}
\usepackage{comment}
\usepackage{booktabs}
\usepackage{soul}
\usepackage{titlesec}
\titleformat{\section}{\normalfont\large\bf\center}{\thesection.}{1em}{}
\titleformat{\subsection}{\normalfont\SmallTitleFont\bf\raggedright}{\thesubsection.}{1em}{}
\titleformat{\subsubsection}{\normalfont\normalsize\bf\raggedright}{\thesubsubsection.}{1em}{}
\renewcommand\thesection{\arabic{section}}
\renewcommand\thesubsection{\thesection.\arabic{subsection}}
\renewcommand\thesubsubsection{\thesubsection.\arabic{subsubsection}}
\usepackage{epstopdf}
\usepackage[latin1]{inputenc}

\usepackage{hyperref}
\usepackage{xstring}

\usepackage{xcolor}

\newcommand\blfootnote[1]{%
  \begingroup
  \renewcommand\thefootnote{}\footnote{#1}%
  \addtocounter{footnote}{-1}%
  \endgroup
}

\title{On the Robustness of Unsupervised and Semi-supervised \\ Cross-lingual Word Embedding Learning}

\name{Yerai Doval*,\textsuperscript{1} Jose Camacho-Collados*,\textsuperscript{2} Luis Espinosa-Anke,\textsuperscript{2} Steven Schockaert\textsuperscript{2}}

\address{\textsuperscript{1}Grupo COLE, Escola Superior de Enxe\~{n}ar\'ia Inform\'atica, Universidade de Vigo, Spain \\ yerai.doval@uvigo.es\\
         \textsuperscript{2}School of Computer Science and Informatics, Cardiff University, UK \\ 
         \{camachocolladosj, espinosa-ankel, schockaerts1\}@cardiff.ac.uk\\}

\abstract{
Cross-lingual word embeddings are vector representations of words in different languages where words with similar meaning are represented by similar vectors, regardless of the language. Recent developments which construct these embeddings by aligning monolingual spaces have shown that accurate alignments can be obtained with little or no supervision, which usually comes in the form of bilingual dictionaries. However, the focus has been on a particular controlled scenario for evaluation, and there is no strong evidence on how current state-of-the-art systems would fare with noisy text or for language pairs with major linguistic differences. In this paper we present an extensive evaluation over multiple cross-lingual embedding models, analyzing their strengths and limitations with respect to different variables such as target language, training corpora and amount of supervision. Our conclusions put in doubt the view that high-quality cross-lingual embeddings can always be learned without much supervision.\\ \newline
}

\begin{document}

\maketitleabstract

\section{Introduction}

The standard approach for training word embeddings is to rely on monolingual corpora, which means in particular that a separate embedding model is learned for each language. 
There is an increasing interest, however, in learning cross-lingual word embeddings, where words from different languages are mapped onto a single space. Such representations are attractive, for instance, for dealing with the multilingual nature of text on the Web, but also as a vehicle for transferring knowledge (e.g., labelled training data) from resource-rich languages such as English to other languages \cite{ruder2017survey}. 
\blfootnote{Authors marked with an asterisk (*) contributed equally.}

Initially, the main obstacle to learning such cross-lingual embeddings was the need for large multilingual parallel corpora \cite{klementiev2012inducing,ap2014autoencoder,luong2015bilingual}. 
This limitation, however, was alleviated by the development of methods that only need comparable data (e.g., Wikipedia corpora in different languages) as the main source of supervision \cite{vulic2015bilingual}. In a complementary direction, it has recently been shown that high-quality cross-lingual embeddings can be obtained by aligning two independently learned monolingual embedding spaces. This strategy is appealing, as it means that one only needs access to monolingual corpora and a bilingual dictionary as supervision signal.
Surprisingly, perhaps, it turns out that dictionaries with less than 100 word pairs are sufficient to obtain good alignments \cite{artetxe-labaka-agirre:2017:Long}. In fact, recent works have shown that cross-lingual embeddings can even be learned without any user-provided dictionary  \cite{conneau2018word,artetxe:acl2018,xu2018crosslingual}.

Despite the promising results reported in the literature, it remains unclear under which conditions the aforementioned methods succeed. For example, \newcite{artetxe-labaka-agirre:2017:Long} and \newcite{conneau2018word} achieved promising results in the word translation task (i.e., bilingual lexicon induction), but their experiments relied on the availability of high-quality monolingual source corpora, namely Wikipedia, which is also the case in a more recent analysis on cross-lingual embeddings performance \cite{glavas2019evaluate}.  
In fact, there exists a significant number of settings which have been largely ignored, and which might challenge models that excel in idealized environments. For instance, \newcite{ahmad2018near} found that for dissimilar languages with different word orderings than English, cross-lingual transfer is still challenging. Similarly, it remains unclear how well existing methods would perform on language pairs with significant differences in morphology (e.g., English-Finnish, the latter being an agglutinative language) or with different alphabets (e.g., English-Farsi or English-Russian). Moreover, settings with different kinds of corpora (e.g. noisy user-generated) have not been fully explored. This means, among others, that it is not clear how current cross-lingual embedding models would behave for transferring knowledge in environments such as social media centred tasks, given that such tasks usually benefit from embeddings that have been trained on social media corpora \cite{tang2014learning,godin2015multimedia,Yang2018}. 

In this work, we broaden the empirical evaluation of state-of-the-art techniques for learning cross-lingual embeddings, by using several types of training corpora, various amounts of supervision, languages from different families and different alignment strategies in three different tasks. The results obtained cast some doubt on the view that high-quality cross-lingual embeddings can always be learned without much supervision. 

\section{Related Work}

\noindent Cross-lingual embeddings have become increasingly popular in the past few years \cite{smith2017offline,artetxe-labaka-agirre:2017:Long,artetxe2018generalizing,conneau2018word}. 
Recent efforts have focused on reducing the need for large amounts of resources (e.g., parallel corpora), which could be difficult to obtain for most languages and language pairs.
However, the evaluation of these approaches has tended to be somewhat limited, often using only one type of training corpus, including only similar languages, and considering only one evaluation task. 
The most similar work to ours is that of \newcite{sogaard2018limitations}, which included an in-depth analysis of two of the factors that we also considered, namely language family and corpus type, but they only considered a single model, i.e., \texttt{MUSE} \cite{conneau2018word}. 
Moreover, they studied each factor in isolation. In our case the analysis is also extended to more languages (covering up to 5 language pairs), systems (two unsupervised, two supervised, and a postprocessing technique), evaluation tasks (cross-lingual word similarity), and the impact of external bilingual dictionaries.

Another similar contribution is the analysis by \newcite{vulic2016role}, where the impact of bilingual dictionaries on cross-lingual alignments was examined. However, they only considered closely-related languages using the same alphabet and one type of corpus (i.e., Wikipedia).
Also, given the publication date, this analysis does not account for the important developments in cross-lingual embeddings from recent years, such as the methods we cover in this paper. Other empirical comparisons focused mostly on the need for different degrees of supervision, such as \cite{upadhyay2016cross}, which has been extended in a more recent survey by \newcite{ruder2017survey}. 

In this paper, we complement those studies by analyzing and discussing empirical findings of the most recent state-of-the-art unsupervised and semi-supervised methods in a broader experimental setting, more in line with the recent concurrent analysis of \newcite{glavas2019evaluate}. The main differences between this empirical evaluation and the contributions of our work lie in the scope of the survey, since: (1) they only consider Wikipedia data for training; (2) they do not consider postprocessing techniques such as \texttt{Meemi}~\cite{doval:meemiemnlp2018}, which we found to improve the performance of cross-lingual models, especially in the case of distant languages and non-comparable corpora; (3) in our analysis we also consider additional settings with scarce training data such as small seed dictionaries and automatically-constructed dictionaries; and (4) we include a more exhaustive intrinsic evaluation (including cross-lingual semantic similarity).

\section{Learning Cross-lingual Word Embeddings}
\label{methods}

The focus of our evaluation is on methods
that start off with monolingual embedding models and then integrate these in a shared cross-lingual space.

Hence, given two monolingual corpora, a word vector space is first learned independently for each language. This can be achieved with common word embedding models, e.g., Word2Vec \cite{Mikolovetal:2013}, GloVe \cite{pennington2014glove} or FastText \cite{bojanowski2017enriching}. Second, a linear alignment strategy is used to map the monolingual embeddings to a common bilingual vector space (Section \ref{alignments}). 
In some cases, a third transformation is applied to already aligned embeddings so the word vectors from both languages are refined and further re-positioned (Section \ref{postprocessing}). 
Regardless of the overall methodology, however, these linear transformations are all learned based on a bilingual dictionary. This dictionary may be manually curated or, in some cases, automatically generated as part of the alignment process.

\subsection{Alignment methods}
\label{alignments}

In this paper we analyze two well-known orthogonal models for aligning monolingual embedding models: the corresponding version of \texttt{VecMap} and \texttt{MUSE}, plus an unsupervised non-orthogonal variant of the former. Basically, both methods use a linear transformation learned through an iterative procedure in which a seed bilingual dictionary is iteratively refined. 
They can be used 
with an empty initial seed dictionary, in which case the alignment process is fully unsupervised.

\texttt{VecMap} \cite{artetxe-labaka-agirre:2017:Long} uses an orthogonal transformation over normalized word embeddings. 
Its semi-supervised two-step procedure is specifically designed to avoid the need for a large seed dictionary. For instance, in the original paper, a seed dictionary with 25 word pairs was used. 
This seed dictionary is then augmented by applying the learned transformation to new words from the source language. The process is repeated until some convergence criterion is met. 
The unsupervised variant \cite{artetxe:acl2018} obtains the initial seed dictionary automatically by exploiting the similarity distribution of words, and then applies the same method followed by a refinement step that re-weights the embeddings based on the cross-correlation of their components, which makes it the only non-orthogonal method tested in this work.
\texttt{MUSE} \cite{conneau2018word} obtains its transformation matrix in a similar way. 
In this case, the seed dictionary is used as-is (supervised setting) or obtained in a fully automatically way through an adversarial learning method (unsupervised setting).

\subsection{Limitations and postprocessing}
\label{postprocessing}

By restricting transformations to orthogonal linear mappings, \texttt{VecMap} and \texttt{MUSE} rely on the assumption that the monolingual embeddings spaces are approximately isomorphic \cite{barone2016towards}. However, it has been argued that this assumption is overly restrictive, as the isomorphism assumption is not always satisfied \cite{sogaard2018limitations,yuva2018generalizing}. For this reason, it has been proposed to go beyond orthogonal transformations by modifying the internal structure of the monolingual spaces, either by giving more weight to highly correlated embedding components, as is the case for the unsupervised variant of \texttt{VecMap} in this work \cite{artetxe2018generalizing}, or by complementing the orthogonal transformation with other forms of post-processing. As an example of this latter strategy, \newcite{doval:meemiemnlp2018} fine-tune the initial alignment by learning an unconstrained linear transformation which aims to map each word vector onto the average of that vector and the corresponding word vector from the other language.

\section{Variables} 

Our main aim is to explore how the choice of corpora (Section \ref{corpora}), supervision signals (Section \ref{supervision}) and languages (Section \ref{languages}) impacts the performance of cross-lingual word embedding models. In Section \ref{others} we also list some other variables which were not directly studied in this paper.

\subsection{Monolingual corpora}
\label{corpora}

It is reasonable to assume that accurate word-level alignments will be easier to obtain from corpora from similar domains with similar vocabularies and register. 
Wikipedia has been the mainstream monolingual source in cross-lingual word embedding training so far \cite{artetxe-labaka-agirre:2017:Long,conneau2018word}. It provides a particularly reliable bilingual signal because of the highly comparative nature of Wikipedia corpora from different languages. As we will see, this makes finding high-quality alignments considerably easier. 

In our analysis we use three different types of corpora: 
Wikipedia\footnote{All Wikipedia text dumps were downloaded from the Polyglot project \cite{polyglot:2013:ACL-CoNLL}: \url{https://sites.google.com/site/rmyeid/projects/polyglot}} (as a prototypical example of comparable monolingual corpora), Web corpora from different sources\footnote{The sources of the web-corpora are: UMBC \cite{han2013umbc}, 1-billion \cite{cardellinoSBWCE}, itWaC and sdeWaC \cite{baroni2009wacky}, Hamshahri \cite{aleahmad2009hamshahri}, and Common Crawl downloaded from \url{http://www.statmt.org/wmt16/translation-task.html}.} (as a prototypical example of non-comparable but generally high-quality corpora) and social media\footnote{Social media corpora are based on Twitter, at different dates between 2015 and 2018 \cite{camacho2020learning}. Monolingual embeddings were downloaded at \url{https://github.com/pedrada88/crossembeddings-twitter}} (as a prototypical example of noisy text). Statistics of these corpora are provided in Table \ref{tab:corpora}.\footnote{Due to some restrictions, we were not able to compile a reliable Twitter corpus for Russian.}

\begin{table}
\begin{center}
{
\setlength{\tabcolsep}{5.0pt}
\scalebox{0.95}{ 
\begin{tabular}{llcrrrcc}
\toprule
\bf Domain & \bf Corpus &   \bf Language & \bf  Size & \bf Words \\
\midrule
\multirow{6}{*}{Wikipedia} &  Wiki\textsubscript{en} & English &  1.7B & 12.0M  \\
 &  Wiki\textsubscript{es} & Spanish &  407M & 3.4M   \\
&  Wiki\textsubscript{it} & Italian &  338M & 3.3M   \\
&  Wiki\textsubscript{de} & German &  605M & 7.4M   \\
  &  Wiki\textsubscript{fi} & Finnish &  68%.6
  M & 2.8M   \\
   &  Wiki\textsubscript{ru} & Russian &  313M & 5.4M   \\
 &  Wiki\textsubscript{fa} & Farsi &  48M & 1.0M   \\

 \midrule

\multirow{6}{*}{\shortstack{Web\\corpora}} & UMBC & English &  %1.18B 
3.5B & %8.07M  
8.1M   \\
% &  1-billion & Spanish &  %1.88B 
% - & - \\
%&  itWaC & Italian & -  & -   \\
%&  sdeWaC & German &  - & -  \\
 % &  Comm. crawl & Finnish &  - & -   \\
 %&  Hamshahri & Farsi & - & -   \\
 
  &  1-billion & Spanish &  %1.88B 
 1.9B & 5.5M \\
&  itWaC & Italian & 1.3B  & 4.2M   \\
&  sdeWaC & German &  438M & %1.48M  
1.5M\\
  &  Comm-crawl & Finnish &  2.8B & %1.78M
  1.8M\\
    &  Comm-crawl & Russian & 1.1B & 18.8M \\
 &  Hamshahri & Farsi & 167M & %0.82M
 0.8M \\
 \midrule

 \multirow{6}{*}{\shortstack{Social\\media}} &  Twitter\textsubscript{en} & English &  294%.3
 M & 5.5M   \\
 &  Twitter\textsubscript{es} & Spanish &  144%.4
 M & 3.3M   \\
&  Twitter\textsubscript{it} & Italian &  63%.1
M & 1.6M   \\
&  Twitter\textsubscript{de} & German &  114%.5
M & 2.3M   \\
 &  Twitter\textsubscript{fi} & Finnish &  29%.0
 M & 1.7M  \\
 &  Twitter\textsubscript{fa} & Farsi &  90%.3
 M & 1.0M   \\

\bottomrule
\end{tabular}
}
}
\end{center}
\caption{\label{tab:corpora} Statistics of the corpora used to train monolingual word embeddings: size (measured in total number of tokens) and words (number of unique tokens).}
\end{table}

\subsection{Bilingual supervision}
\label{supervision}

Early approaches for learning bilingual embeddings relied on large parallel corpora \cite{klementiev2012inducing,luong2015bilingual}, which limited their applicability. More recent approaches instead rely on (often small) bilingual dictionaries as the only source of bilingual supervision.  
In fact, some methods remove the need for a user-supplied bilingual dictionary altogether \cite{conneau2018word,artetxe:acl2018,hoshen-wolf-2018-non,xu2018crosslingual}, relying instead on synthetic dictionaries that are obtained fully automatically. 
In our experiments we consider a wide range of signals, including no supervision as well as automatically generated dictionaries of identical words. In the latter case, we rely on the assumption that words that occur in both of the monolingual corpora tend to have the same meaning. While this may seem naive, this strategy has been reported in the literature to perform well in practice \cite{smith2017offline,sogaard2018limitations}. 

\subsection{Languages}
\label{languages}

In most previous work, the evaluation of cross-lingual embeddings has been limited to a small set of closely-related languages. For instance, \newcite{smith2017offline} evaluated their model on the English-Italian pair only, while the evaluation of \newcite{artetxe-labaka-agirre:2017:Long} was performed on three languages, all of which share the same alphabet.  Moreover, as the considered language pairs vary from one study to another, the relative performance of different methods for particular types of languages remains unclear.
More recently, however, \newcite{sogaard2018limitations} have extended the usual evaluation framework by covering additional Eastern European languages.  
We similarly expand the range of languages by considering: Spanish (ES), Italian (IT), German (DE), Finnish (FI), Farsi (FA) and Russian (RU). In all cases we use English (EN) as source language. This set of languages represents not only the usual family of Indo-European languages (all of them except Finnish), but also agglutinative languages (German, Farsi and Finnish, the latter being non-Indo-European), as well as languages with different alphabets (Farsi and Russian). 

\begin{table*}[!t]
\renewcommand{\arraystretch}{1.15}
\resizebox{\textwidth}{!}{
\begin{tabular}{llrrrrrrrrrrrrrrrrrrr}
\hline
\multicolumn{21}{c}{\textbf{\texttt{Wikipedia}}} \\ \hline
\multicolumn{1}{l|}{\multirow{2}{*}{\textbf{Sup.}}} & \multicolumn{1}{l|}{\multirow{2}{*}{\textbf{Model}}} & \multicolumn{3}{c|}{\textbf{Spanish}} & \multicolumn{3}{c|}{\textbf{Italian}} & \multicolumn{3}{c|}{\textbf{German}} & \multicolumn{3}{c|}{\textbf{Finnish}} & \multicolumn{3}{c|}{\textbf{Farsi}} & \multicolumn{3}{c|}{\textbf{Russian}} &
\multicolumn{1}{c}{\textbf{Avg}} \\ \cline{3-21} 
\multicolumn{1}{l|}{} & \multicolumn{1}{l|}{} & \multicolumn{1}{l}{$P@1$} & \multicolumn{1}{l}{$P@5$} & \multicolumn{1}{l|}{$P@10$} & \multicolumn{1}{l}{$P@1$} & \multicolumn{1}{l}{$P@5$} & \multicolumn{1}{l|}{$P@10$} & \multicolumn{1}{l}{$P@1$} & \multicolumn{1}{l}{$P@5$} & \multicolumn{1}{l|}{$P@10$} & \multicolumn{1}{l}{$P@1$} & \multicolumn{1}{l}{$P@5$} & \multicolumn{1}{l|}{$P@10$} & \multicolumn{1}{l}{$P@1$} & \multicolumn{1}{l}{$P@5$} & \multicolumn{1}{l|}{$P@10$}  & \multicolumn{1}{l}{$P@1$} & \multicolumn{1}{l}{$P@5$} & \multicolumn{1}{l|}{$P@10$} & \multicolumn{1}{c}{$P@5$} \\ \hline
\multicolumn{1}{l|}{\multirow{2}{*}{\textbf{\texttt{Unsup}}}} & \multicolumn{1}{l|}{\textbf{VecMap}}  & \textbf{39.6} & 66.1 & \multicolumn{1}{r|}{72.3} & 42.7 & 65.7 & \multicolumn{1}{r|}{71.6} & \textbf{28.6} & \textbf{48.3} & \multicolumn{1}{r|}{\textbf{54.8}} & 19.6 & 40.4 & \multicolumn{1}{r|}{48.3} & 20.5 & 37.0 & \multicolumn{1}{r|}{42.8} & 19.5 & 45.3 & \multicolumn{1}{r|}{54.5} & 50.5 \\
\multicolumn{1}{l|}{} & \multicolumn{1}{l|}{\textbf{MUSE}} & 39.3 & 64.7 & \multicolumn{1}{r|}{71.3} & 41.6 & 63.2 & \multicolumn{1}{r|}{69.9} & 28.3 & 46.5 & \multicolumn{1}{r|}{53.3} & 0.0 & 0.0 & \multicolumn{1}{r|}{0.0} & 0.0 & 0.0 & \multicolumn{1}{r|}{0.0} & 14.9 & 36.0 & \multicolumn{1}{r|}{46.5} & 35.1 \\
  \hline
  
\multicolumn{1}{l|}{\multirow{2}{*}{\textbf{\texttt{Ident}}}} & \multicolumn{1}{l|}{\textbf{VecMap}}  & \textbf{39.5} & 66.0 & \multicolumn{1}{r|}{72.4} & 42.7 & 65.8 & \multicolumn{1}{r|}{71.7} & \textbf{28.6} & \textbf{48.3} & \multicolumn{1}{r|}{54.7} & 21.6 & 43.7 & \multicolumn{1}{r|}{51.6} & 23.4 & 40.3 & \multicolumn{1}{r|}{46.1} & 19.6 & 46.0 & \multicolumn{1}{r|}{55.1} & 51.7 \\
\multicolumn{1}{l|}{} & \multicolumn{1}{l|}{\textbf{MUSE}} & 35.9 & 60.6 & \multicolumn{1}{r|}{67.3} & 37.8 & 60.4 & \multicolumn{1}{r|}{68.5} & 24.8 & 41.9 & \multicolumn{1}{r|}{49.5} & 13.4 & 25.5 & \multicolumn{1}{r|}{32.0} & 6.7 & 16.6 & \multicolumn{1}{r|}{21.3} & 7.8 & 19.9 & \multicolumn{1}{r|}{26.1} & 37.5  \\
  \hline
  
\multicolumn{1}{l|}{\multirow{4}{*}{\textbf{\texttt{8K}}}} & \multicolumn{1}{l|}{\textbf{VecMap}} & \textbf{39.6} & 66.2 & \multicolumn{1}{r|}{72.3} & \textbf{42.6} & 65.9 & \multicolumn{1}{r|}{71.8} & \textbf{28.6} & \textbf{48.3} & \multicolumn{1}{r|}{\textbf{54.8}} & 22.4 & 44.5 & \multicolumn{1}{r|}{52.5} & 22.8 & 39.7 & \multicolumn{1}{r|}{46.2} & 20.0 & 46.3 & \multicolumn{1}{r|}{55.6} & 51.8  \\
\multicolumn{1}{l|}{} & \multicolumn{1}{l|}{\textbf{MUSE}} & 39.1 & 65.4 & \multicolumn{1}{r|}{72.3} & 41.1 & 63.3 & \multicolumn{1}{r|}{70.1} & 27.6 & 45.9 & \multicolumn{1}{r|}{53.2} & 19.5 & 40.4 & \multicolumn{1}{r|}{49.5} & 19.7 & 35.4 & \multicolumn{1}{r|}{42} & 21.3 & 43.7 & \multicolumn{1}{r|}{52.9} & 49.0  \\
\multicolumn{1}{l|}{} & \multicolumn{1}{l|}{\textbf{Meemi}\textsubscript{VM}}  & 39.3 & \textbf{67.4} & \multicolumn{1}{r|}{\textbf{73.7}} & 41.6 & 66.5 & \multicolumn{1}{r|}{72.5} & 28 & 47.8 & \multicolumn{1}{r|}{\textbf{54.8}} & \textbf{23.8} & \textbf{48.7} & \multicolumn{1}{r|}{\textbf{57.0}} & \textbf{23.4} & \textbf{41.7} & \multicolumn{1}{r|}{\textbf{47.7}} & 23.0 & 49.3 & \multicolumn{1}{r|}{58.3} & \textbf{53.4}  \\
\multicolumn{1}{l|}{} & \multicolumn{1}{l|}{\textbf{Meemi}\textsubscript{MS}} & 39.3 & \textbf{67.4} & \multicolumn{1}{r|}{\textbf{73.7}} & 41.3 & \textbf{66.8} & \multicolumn{1}{r|}{\textbf{72.8}} & 27.1 & 46.3 & \multicolumn{1}{r|}{53.9} & 21.7 & 45.0 & \multicolumn{1}{r|}{53.6} & 20.7 & 38.6 & \multicolumn{1}{r|}{45.1} & \textbf{24.4} & \textbf{50.3} & \multicolumn{1}{r|}{\textbf{59.3}} & 52.4  \\
 \hline

\multicolumn{21}{c}{\textbf{\texttt{Web corpora}}} \\ \hline
\multicolumn{1}{l|}{\multirow{2}{*}{\textbf{Sup.}}} & \multicolumn{1}{l|}{\multirow{2}{*}{\textbf{Model}}} & \multicolumn{3}{c|}{\textbf{Spanish}} & \multicolumn{3}{c|}{\textbf{Italian}} & \multicolumn{3}{c|}{\textbf{German}} & \multicolumn{3}{c|}{\textbf{Finnish}} & \multicolumn{3}{c|}{\textbf{Farsi}} & \multicolumn{3}{c|}{\textbf{Russian}} &
\multicolumn{1}{c}{\textbf{Avg}} \\ \cline{3-21} 
\multicolumn{1}{l|}{} & \multicolumn{1}{l|}{} & \multicolumn{1}{l}{$P@1$} & \multicolumn{1}{l}{$P@5$} & \multicolumn{1}{l|}{$P@10$} & \multicolumn{1}{l}{$P@1$} & \multicolumn{1}{l}{$P@5$} & \multicolumn{1}{l|}{$P@10$} & \multicolumn{1}{l}{$P@1$} & \multicolumn{1}{l}{$P@5$} & \multicolumn{1}{l|}{$P@10$} & \multicolumn{1}{l}{$P@1$} & \multicolumn{1}{l}{$P@5$} & \multicolumn{1}{l|}{$P@10$} & \multicolumn{1}{l}{$P@1$} & \multicolumn{1}{l}{$P@5$} & \multicolumn{1}{l|}{$P@10$}  & \multicolumn{1}{l}{$P@1$} & \multicolumn{1}{l}{$P@5$} & \multicolumn{1}{l|}{$P@10$} & \multicolumn{1}{c}{$P@5$} \\ \hline
\multicolumn{1}{l|}{\multirow{2}{*}{\textbf{\texttt{Unsup}}}} & \multicolumn{1}{l|}{\textbf{VecMap}}   & \textbf{34.8} & 60.6 & \multicolumn{1}{r|}{67.0} & 31.4 & 53.7 & \multicolumn{1}{r|}{60.7} & 23.2 & 42.7 & \multicolumn{1}{r|}{50.2} & 0.0 & 0.0 & \multicolumn{1}{r|}{0.0} & 19.7 & 34.6 & \multicolumn{1}{r|}{40.4} & 13.8 & 30.9 & \multicolumn{1}{r|}{38.6} & 37.1  \\
\multicolumn{1}{l|}{} & \multicolumn{1}{l|}{\textbf{MUSE}}  & 31.4 & 51.2 & \multicolumn{1}{r|}{57.7} & 31.4 & 51.2 & \multicolumn{1}{r|}{57.7} & 20.8 & 38.7 & \multicolumn{1}{r|}{46.6} & 17.7 & 35.7 & \multicolumn{1}{r|}{42.8} & 18.1 & 32.8 & \multicolumn{1}{r|}{37.8} & 0.0 & 0.0 & \multicolumn{1}{r|}{0.0} & 34.9  \\
  \hline
  
\multicolumn{1}{l|}{\multirow{2}{*}{\textbf{\texttt{Ident}}}} & \multicolumn{1}{l|}{\textbf{VecMap}}   & 34.7 & 60.4 & \multicolumn{1}{r|}{67.0} & 31.4 & 54.0 & \multicolumn{1}{r|}{60.7} & 23.1 & 42.9 & \multicolumn{1}{r|}{50.5} & 18.6 & 41.6 & \multicolumn{1}{r|}{49.3} & 20.0 & 35.3 & \multicolumn{1}{r|}{40.3} & 14.1 & 31.2 & \multicolumn{1}{r|}{38.8} & 44.2  \\
\multicolumn{1}{l|}{} & \multicolumn{1}{l|}{\textbf{MUSE}}  & 26.1 & 46.7 & \multicolumn{1}{r|}{53.8} & 24.7 & 45.1 & \multicolumn{1}{r|}{52.4} & 17.4 & 32.8 & \multicolumn{1}{r|}{40.5} & 12.6 & 26.0 & \multicolumn{1}{r|}{33.8} & 3.0 & 8.3 & \multicolumn{1}{r|}{5.8} & 0.1 & 0.2 & \multicolumn{1}{r|}{0.2} & 26.5  \\
  \hline
  
\multicolumn{1}{l|}{\multirow{4}{*}{\textbf{\texttt{8K}}}} & \multicolumn{1}{l|}{\textbf{VecMap}} & 34.6 & 60.6 & \multicolumn{1}{r|}{66.9} & 31.9 & 54.2 & \multicolumn{1}{r|}{60.4} & 23.1 & 42.7 & \multicolumn{1}{r|}{50.5} & 18.9 & 40.9 & \multicolumn{1}{r|}{48.8} & 19.6 & 35.8 & \multicolumn{1}{r|}{41.4} & 14.6 & 31.7 & \multicolumn{1}{r|}{39.6} & 44.3  \\
\multicolumn{1}{l|}{} & \multicolumn{1}{l|}{\textbf{MUSE}}  &  32.5 & 58.2 & \multicolumn{1}{r|}{65.9} & 32.5 & 56.0 & \multicolumn{1}{r|}{63.2} & 22.4 & 40.9 & \multicolumn{1}{r|}{48.9} & 20.0 & 40.1 & \multicolumn{1}{r|}{48.3} & 17.4 & 31.6 & \multicolumn{1}{r|}{37.6} & 15.5 & 35.6 & \multicolumn{1}{r|}{44.1} & 43.7  \\
\multicolumn{1}{l|}{} & \multicolumn{1}{l|}{\textbf{Meemi}\textsubscript{VM}} &  34.5 & \textbf{61.6} & \multicolumn{1}{r|}{67.9} & 33.6 & 58.3 & \multicolumn{1}{r|}{\textbf{65.6}} & \textbf{23.7} & \textbf{45.4} & \multicolumn{1}{r|}{\textbf{53.2}} & 22.3 & \textbf{46.7} & \multicolumn{1}{r|}{\textbf{55.0}} & \textbf{21.7} & \textbf{39.0} & \multicolumn{1}{r|}{\textbf{43.8}} & 18.2 & 40.0 & \multicolumn{1}{r|}{47.5} & \textbf{48.9}  \\
\multicolumn{1}{l|}{} & \multicolumn{1}{l|}{\textbf{Meemi}\textsubscript{MS}}  & 33.9 & 60.7 & \multicolumn{1}{r|}{\textbf{68.4}} & \textbf{33.8} & \textbf{58.4} & \multicolumn{1}{r|}{\textbf{65.6}} & \textbf{23.7} & 45.3 & \multicolumn{1}{r|}{52.3} & \textbf{23.0} & 46.1 & \multicolumn{1}{r|}{54.0} & 19.3 & 36.0 & \multicolumn{1}{r|}{41.7} & \textbf{18.7} & \textbf{40.5} & \multicolumn{1}{r|}{\textbf{49.7}} & 47.8  \\
  \hline

\multicolumn{18}{c}{\textbf{\texttt{Social media}}} \\ \cline{1-18}
\multicolumn{1}{l|}{\multirow{2}{*}{\textbf{Sup.}}} & \multicolumn{1}{l|}{\multirow{2}{*}{\textbf{Model}}} & \multicolumn{3}{c|}{\textbf{Spanish}} & \multicolumn{3}{c|}{\textbf{Italian}} & \multicolumn{3}{c|}{\textbf{German}} & \multicolumn{3}{c|}{\textbf{Finnish}} & \multicolumn{3}{c|}{\textbf{Farsi}} &
\multicolumn{1}{c}{\textbf{Avg}} \\ \cline{3-18} 
\multicolumn{1}{l|}{} & \multicolumn{1}{l|}{} & \multicolumn{1}{l}{$P@1$} & \multicolumn{1}{l}{$P@5$} & \multicolumn{1}{l|}{$P@10$} & \multicolumn{1}{l}{$P@1$} & \multicolumn{1}{l}{$P@5$} & \multicolumn{1}{l|}{$P@10$} & \multicolumn{1}{l}{$P@1$} & \multicolumn{1}{l}{$P@5$} & \multicolumn{1}{l|}{$P@10$} & \multicolumn{1}{l}{$P@1$} & \multicolumn{1}{l}{$P@5$} & \multicolumn{1}{l|}{$P@10$} & \multicolumn{1}{l}{$P@1$} & \multicolumn{1}{l}{$P@5$} & \multicolumn{1}{l|}{$P@10$} & \multicolumn{1}{c}{$P@5$} \\ \cline{1-18}
\multicolumn{1}{l|}{\multirow{2}{*}{\textbf{\texttt{Unsup}}}} & \multicolumn{1}{l|}{\textbf{VecMap}}   & \multicolumn{1}{r}{8.1} & \multicolumn{1}{r}{16.4} & \multicolumn{1}{r|}{20.4} & \multicolumn{1}{r}{8.8} & \multicolumn{1}{r}{17.0} & \multicolumn{1}{r|}{22.3} & \multicolumn{1}{r}{0.1} & \multicolumn{1}{r}{0.4} & \multicolumn{1}{r|}{0.5} & \multicolumn{1}{r}{0.0} & \multicolumn{1}{r}{0.0} & \multicolumn{1}{r|}{0.0} & \multicolumn{1}{r}{0.0} & \multicolumn{1}{r}{0.0} & \multicolumn{1}{r|}{0.0} & \multicolumn{1}{r}{6.8} \\
\multicolumn{1}{l|}{} & \multicolumn{1}{l|}{\textbf{MUSE}}   & \multicolumn{1}{r}{0.0} & \multicolumn{1}{r}{0.0} & \multicolumn{1}{r|}{0.0} & \multicolumn{1}{r}{7.3} & \multicolumn{1}{r}{14.5} & \multicolumn{1}{r|}{18.3} & \multicolumn{1}{r}{0.0} & \multicolumn{1}{r}{0.0} & \multicolumn{1}{r|}{0.0} & \multicolumn{1}{r}{0.0} & \multicolumn{1}{r}{0.0} & \multicolumn{1}{r|}{0.1} & \multicolumn{1}{r}{0.0} & \multicolumn{1}{r}{0.1} & \multicolumn{1}{r|}{0.1} & \multicolumn{1}{r}{2.9} \\
  \cline{1-18}
  
\multicolumn{1}{l|}{\multirow{2}{*}{\textbf{\texttt{Ident}}}} & \multicolumn{1}{l|}{\textbf{VecMap}}   & \multicolumn{1}{r}{8.1} & \multicolumn{1}{r}{16.4} & \multicolumn{1}{r|}{20.4} & \multicolumn{1}{r}{8.8} & \multicolumn{1}{r}{17.0} & \multicolumn{1}{r|}{22.3} & \multicolumn{1}{r}{0.1} & \multicolumn{1}{r}{0.4} & \multicolumn{1}{r|}{0.5} & \multicolumn{1}{r}{0.0} & \multicolumn{1}{r}{0.0} & \multicolumn{1}{r|}{0.0} & \multicolumn{1}{r}{0.0} & \multicolumn{1}{r}{0.0} & \multicolumn{1}{r|}{0.0} & \multicolumn{1}{r}{8.2} \\
\multicolumn{1}{l|}{} & \multicolumn{1}{l|}{\textbf{MUSE}}   & \multicolumn{1}{r}{0.0} & \multicolumn{1}{r}{0.0} & \multicolumn{1}{r|}{0.0} & \multicolumn{1}{r}{7.3} & \multicolumn{1}{r}{14.5} & \multicolumn{1}{r|}{18.3} & \multicolumn{1}{r}{0.0} & \multicolumn{1}{r}{0.0} & \multicolumn{1}{r|}{0.0} & \multicolumn{1}{r}{0.0} & \multicolumn{1}{r}{0.0} & \multicolumn{1}{r|}{0.1} & \multicolumn{1}{r}{0.0} & \multicolumn{1}{r}{0.1} & \multicolumn{1}{r|}{0.1} & \multicolumn{1}{r}{3.7} \\
  \cline{1-18}
  
\multicolumn{1}{l|}{\multirow{4}{*}{\textbf{\texttt{8K}}}}  & \multicolumn{1}{l|}{\textbf{VecMap}} & \multicolumn{1}{r}{8.7} & \multicolumn{1}{r}{16.6} & \multicolumn{1}{r|}{21.6} & \multicolumn{1}{r}{8.9} & \multicolumn{1}{r}{17.3} & \multicolumn{1}{r|}{22.4} & \multicolumn{1}{r}{3.2} & \multicolumn{1}{r}{6.8} & \multicolumn{1}{r|}{9.5} & \multicolumn{1}{r}{0.2} & \multicolumn{1}{r}{0.8} & \multicolumn{1}{r|}{1.2} & \multicolumn{1}{r}{0.4} & \multicolumn{1}{r}{1.6} & \multicolumn{1}{r|}{2.0} & \multicolumn{1}{r}{8.6} \\
\multicolumn{1}{l|}{} & \multicolumn{1}{l|}{\textbf{MUSE}}  & \multicolumn{1}{r}{8.1} & \multicolumn{1}{r}{17.6} & \multicolumn{1}{r|}{22.7} & \multicolumn{1}{r}{8} & \multicolumn{1}{r}{16.4} & \multicolumn{1}{r|}{21.1} & \multicolumn{1}{r}{2.2} & \multicolumn{1}{r}{6.0} & \multicolumn{1}{r|}{8.4} & \multicolumn{1}{r}{0.6} & \multicolumn{1}{r}{2.2} & \multicolumn{1}{r|}{3.2} & \multicolumn{1}{r}{1.2} & \multicolumn{1}{r}{4.5} & \multicolumn{1}{r|}{6.3} & \multicolumn{1}{r}{9.3} \\
\multicolumn{1}{l|}{} & \multicolumn{1}{l|}{\textbf{Meemi}\textsubscript{VM}}  & \multicolumn{1}{r}{\textbf{9.8}} & \multicolumn{1}{r}{\textbf{21.3}} & \multicolumn{1}{r|}{\textbf{26.9}} & \multicolumn{1}{r}{\textbf{10.6}} & \multicolumn{1}{r}{\textbf{20.0}} & \multicolumn{1}{r|}{\textbf{25.6}} & \multicolumn{1}{r}{\textbf{3.7}} & \multicolumn{1}{r}{\textbf{9.6}} & \multicolumn{1}{r|}{\textbf{13.2}} & \multicolumn{1}{r}{1.3} & \multicolumn{1}{r}{3.6} & \multicolumn{1}{r|}{5.5} & \multicolumn{1}{r}{\textbf{1.8}} & \multicolumn{1}{r}{5.1} & \multicolumn{1}{r|}{7.0} & \multicolumn{1}{r}{\textbf{12.1}} \\
\multicolumn{1}{l|}{} & \multicolumn{1}{l|}{\textbf{Meemi}\textsubscript{MS}} & \multicolumn{1}{r}{9.5} & \multicolumn{1}{r}{20.5} & \multicolumn{1}{r|}{26.3} & \multicolumn{1}{r}{9.5} & \multicolumn{1}{r}{19.1} & \multicolumn{1}{r|}{24.5} & \multicolumn{1}{r}{3.0} & \multicolumn{1}{r}{7.6} & \multicolumn{1}{r|}{11.1} & \multicolumn{1}{r}{\textbf{1.5}} & \multicolumn{1}{r}{\textbf{4.3}} & \multicolumn{1}{r|}{\textbf{6.4}} & \multicolumn{1}{r}{1.6} & \multicolumn{1}{r}{\textbf{5.3}} & \multicolumn{1}{r|}{\textbf{8.1}} & \multicolumn{1}{r}{11.4} \\
 \cline{1-18}

\end{tabular}
}
\caption{Bilingual dictionary induction results using English as source language. Performance measured by $P@k$. Overall average $P@5$ is shown in the last column.} 
\label{tab:dictionary-r}
\end{table*}

\subsection{Other variables}
\label{others}

It is worth mentioning that there are several other external factors that may affect the quality of cross-lingual embeddings, beyond the ones considered in this study. 
For instance, in our experiments we use FastText \cite{bojanowski2017enriching}, since morphological information might be useful for agglutinative languages as noted by its authors, with default values and dimensionality\footnote{300 dimensions in the case of Wikipedia and web corpora, and 100 dimensions in the smaller social media corpora.}, but the impact of other word embedding models such as Word2Vec \cite{Mikolovetal:2013} or GloVe \cite{pennington2014glove} could also be analyzed, in the line of \newcite{sogaard2018limitations}. 
Likewise, all cross-lingual models and post-processing technique we evaluate are used \textit{as is}, with their default configurations.

\section{Evaluation}

We use two standard tasks for evaluating cross-lingual word embeddings: bilingual dictionary induction (Section \ref{induction}) and cross-lingual word similarity (Section \ref{similarity}). In addition, we also consider a downstream application: cross-lingual natural language inference (Section \ref{sec:xli}). 

The systems we compare are two well-known cross-lingual embedding methods which can be used in unsupervised and semi-supervised settings, namely the orthogonal version of  \texttt{VecMap}\footnote{\url{https://github.com/artetxem/vecmap}} \cite{artetxe:acl2018} and \texttt{MUSE}\footnote{\url{https://github.com/facebookresearch/MUSE}} \cite{conneau2018word}. 
As seed dictionaries we consider three samples of varying sizes, considering 8K, 1K and 100 word pairs, to test the robustness of the models regarding the amount of supervision available.\footnote{These dictionaries were obtained by splitting the training dictionaries provided by \newcite{conneau2018word}}  For the sake of clarity, in this section we only present results for the largest dictionary (i.e., with 8K word pairs). The results for all the other dictionary sizes are included in the appendix (these results are also considered in the analysis in Section \ref{analysis}).
Additionally, we also leverage synthetic dictionaries, consisting of identical words that are found in the corpora for both languages.
Lastly, using those same bilingual dictionaries, we apply the postprocessing proposed in  \newcite{doval:meemiemnlp2018}\footnote{\url{https://github.com/yeraidm/meemi}} to refine the cross-lingual embeddings obtained by \texttt{VecMap} and \texttt{MUSE}. We will refer to these postprocessed vectors as \texttt{Meemi}\textsubscript{VM} and \texttt{Meemi}\textsubscript{MS}, respectively. 

\subsection{Bilingual dictionary induction}
\label{induction}

This task consists in automatically obtaining the word translations in a target language for words in a source language.
To obtain the translation candidates, we use the standard cosine distance measure, selecting the nearest neighbors from the target language to the source word in the cross-lingual embedding space. 
The performance is measured with precision at $k$ ($P@k$), that is, the proportion of test instances where the correct translation candidate for a given source word was among the $k$ highest ranked candidates. 
Table \ref{tab:dictionary-r} summarizes the results obtained by all comparison systems on the test dictionaries published by \newcite{conneau2018word}. Note that the test dictionaries do not overlap with the dictionaries used for training.

\begin{table}[!t]
%\renewcommand{\arraystretch}{1.02}
%\scalebox{0.75}{
\resizebox{\columnwidth}{!}{
%\small
\begin{tabular}{clrrrrr}
\hline

\multicolumn{7}{c}{\texttt{\textbf{Wikipedia}}} \\ \hline \hline
\multicolumn{1}{c|}{\multirow{1}{*}{\textbf{Sup.}}} & \multicolumn{1}{c|}{\multirow{1}{*}{\textbf{Model}}} & \multicolumn{1}{c|}{\textbf{EN-ES}} & \multicolumn{1}{c|}{\textbf{EN-IT}} & \multicolumn{1}{c|}{\textbf{EN-DE}} & \multicolumn{1}{c||}{\textbf{EN-FA}} & \multicolumn{1}{c}{\textbf{Avg}} \\ \hline
\multicolumn{1}{c|}{\multirow{2}{*}{\textbf{\texttt{Unsup}}}} & \multicolumn{1}{l|}{\textbf{VecMap}} & \multicolumn{1}{r|}{72.1}  & \multicolumn{1}{r|}{70.6} & \multicolumn{1}{r|}{69.3}  & \multicolumn{1}{r||}{61.3} & 68.3 \\
\multicolumn{1}{c|}{} & \multicolumn{1}{l|}{\textbf{MUSE}}  & \multicolumn{1}{r|}{72.6}  & \multicolumn{1}{r|}{71.2}  & \multicolumn{1}{r|}{68.9} & \multicolumn{1}{r||}{6.5} & 54.8  \\
\hline
\multicolumn{1}{c|}{\multirow{2}{*}{\textbf{\texttt{Ident}}}} & \multicolumn{1}{l|}{\textbf{VecMap}} & \multicolumn{1}{r|}{71.8}  & \multicolumn{1}{r|}{70.6} & \multicolumn{1}{r|}{69.3}  & \multicolumn{1}{r||}{61.9}  &  68.4  \\
\multicolumn{1}{c|}{} & \multicolumn{1}{l|}{\textbf{MUSE}}  & \multicolumn{1}{r|}{71.9}  & \multicolumn{1}{r|}{70.5}  & \multicolumn{1}{r|}{68.4} & \multicolumn{1}{r||}{51.3} &  65.5 \\
 \hline
\multicolumn{1}{c|}{\multirow{4}{*}{\textbf{\texttt{8K}}}} & \multicolumn{1}{l|}{\textbf{VecMap}}  & \multicolumn{1}{r|}{71.8} & \multicolumn{1}{r|}{70.6} & \multicolumn{1}{r|}{69.3} & \multicolumn{1}{r||}{61.7} & 68.4  \\
\multicolumn{1}{c|}{} & \multicolumn{1}{l|}{\textbf{MUSE}} & \multicolumn{1}{r|}{72.6}  & \multicolumn{1}{r|}{70.9}  & \multicolumn{1}{r|}{68.9} & \multicolumn{1}{r||}{58.7} & 67.8  \\
\multicolumn{1}{c|}{} & \multicolumn{1}{l|}{\textbf{Meemi}\textsubscript{VM}}  & \multicolumn{1}{r|}{71.9} & \multicolumn{1}{r|}{70.9} & \multicolumn{1}{r|}{\textbf{70.3}} & \multicolumn{1}{r||}{\textbf{63.4}} & 69.1  \\
\multicolumn{1}{c|}{} & \multicolumn{1}{l|}{\textbf{Meemi}\textsubscript{MS}} & \multicolumn{1}{r|}{\textbf{72.9}}  & \multicolumn{1}{r|}{\textbf{71.9}} & \multicolumn{1}{r|}{70.1}  & \multicolumn{1}{r||}{62.0}  & \textbf{69.2}   \\ \hline \hline

\multicolumn{7}{c}{\texttt{\textbf{Web corpora}}} \\ \hline \hline
\multicolumn{1}{c|}{\multirow{1}{*}{\textbf{Sup.}}} & \multicolumn{1}{l|}{\multirow{1}{*}{\textbf{Model}}} & \multicolumn{1}{c|}{\textbf{EN-ES}} & \multicolumn{1}{c|}{\textbf{EN-IT}} & \multicolumn{1}{c|}{\textbf{EN-DE}} & \multicolumn{1}{c||}{\textbf{EN-FA}} & \multicolumn{1}{c}{\textbf{Avg}}  \\ \hline

\multicolumn{1}{c|}{\multirow{2}{*}{\textbf{\texttt{Unsup}}}} & \multicolumn{1}{l|}{\textbf{VecMap}} & \multicolumn{1}{r|}{70.5}  & \multicolumn{1}{r|}{68.8} & \multicolumn{1}{r|}{70.4}  & \multicolumn{1}{r||}{33.4}   & 60.8   \\
\multicolumn{1}{c|}{} & \multicolumn{1}{l|}{\textbf{MUSE}}   &\multicolumn{1}{r|}{71.6} 	 & \multicolumn{1}{r|}{69.4}  & \multicolumn{1}{r|}{70.0} & \multicolumn{1}{r||}{23.8}  & 58.7   \\
\hline

\multicolumn{1}{c|}{\multirow{2}{*}{\textbf{\texttt{Ident}}}} & \multicolumn{1}{l|}{\textbf{VecMap}} & \multicolumn{1}{r|}{70.6}  & \multicolumn{1}{r|}{68.8} & \multicolumn{1}{r|}{70.4}  & \multicolumn{1}{r||}{33.0}  & 60.7   \\
\multicolumn{1}{c|}{} & \multicolumn{1}{l|}{\textbf{MUSE}}  & \multicolumn{1}{r|}{70.1}  & \multicolumn{1}{r|}{67.5}  & \multicolumn{1}{r|}{69.7} & \multicolumn{1}{r||}{14.5}  & 55.5   \\
 \hline
\multicolumn{1}{c|}{\multirow{4}{*}{\textbf{\texttt{8K}}}} & \multicolumn{1}{l|}{\textbf{VecMap}}  & \multicolumn{1}{r|}{70.6} &  \multicolumn{1}{r|}{68.8}  & \multicolumn{1}{r|}{70.4}  & \multicolumn{1}{r||}{\textbf{33.5}}  & 60.8   \\
\multicolumn{1}{c|}{} & \multicolumn{1}{l|}{\textbf{MUSE}}  & \multicolumn{1}{r|}{71.9} & \multicolumn{1}{r|}{70.4}  & \multicolumn{1}{r|}{70.2}  & \multicolumn{1}{r||}{23.9}  & 59.1   \\
\multicolumn{1}{c|}{} & \multicolumn{1}{l|}{\textbf{Meemi}\textsubscript{VM}}  & \multicolumn{1}{r|}{70.9} & \multicolumn{1}{r|}{70.0}  & \multicolumn{1}{r|}{71.8}  & \multicolumn{1}{r||}{\textbf{39.0}}  & \textbf{62.9}   \\
\multicolumn{1}{c|}{} & \multicolumn{1}{l|}{\textbf{Meemi}\textsubscript{MS}}  & \multicolumn{1}{r|}{\textbf{72.3}}  & \multicolumn{1}{r|}{\textbf{71.1}}  & \multicolumn{1}{r|}{\textbf{72.1}} & \multicolumn{1}{r||}{33.0}  & 62.1   \\ \hline \hline

\multicolumn{7}{c}{\texttt{\textbf{Social media}}} \\ \hline \hline
\multicolumn{1}{c|}{\multirow{1}{*}{\textbf{Sup.}}} & \multicolumn{1}{l|}{\multirow{1}{*}{\textbf{Model}}} & \multicolumn{1}{c|}{\textbf{EN-ES}} & \multicolumn{1}{c|}{\textbf{EN-IT}} & \multicolumn{1}{c|}{\textbf{EN-DE}} & \multicolumn{1}{c||}{\textbf{EN-FA}} & \multicolumn{1}{c}{\textbf{Avg}}  \\ \hline

\multicolumn{1}{c|}{\multirow{2}{*}{\textbf{\texttt{Unsup}}}} & \multicolumn{1}{l|}{\textbf{VecMap}} & \multicolumn{1}{r|}{46.9} & \multicolumn{1}{r|}{51.5} & \multicolumn{1}{r|}{31.2} & \multicolumn{1}{r||}{2.4}  & 33.0   \\ 
\multicolumn{1}{c|}{} & \multicolumn{1}{l|}{\textbf{MUSE}}  & \multicolumn{1}{r|}{10.9} & \multicolumn{1}{r|}{49.7} & \multicolumn{1}{r|}{13.0}  & \multicolumn{1}{r||}{4.7}  & 19.6   \\ 
\hline
\multicolumn{1}{c|}{\multirow{2}{*}{\textbf{\texttt{Ident}}}} & \multicolumn{1}{l|}{\textbf{VecMap}} & \multicolumn{1}{r|}{47.1}  & \multicolumn{1}{r|}{51.9} & \multicolumn{1}{r|}{50.3}  & \multicolumn{1}{r||}{26.5}  & 44.0   \\
\multicolumn{1}{c|}{} & \multicolumn{1}{l|}{\textbf{MUSE}}  & \multicolumn{1}{r|}{47.7}  & \multicolumn{1}{r|}{49.8}  & \multicolumn{1}{r|}{46.8} & \multicolumn{1}{r||}{32.4}  & 44.2   \\
\hline
\multicolumn{1}{c|}{\multirow{4}{*}{\textbf{\texttt{8K}}}} & \multicolumn{1}{l|}{\textbf{VecMap}} & \multicolumn{1}{r|}{47.4}  & \multicolumn{1}{r|}{51.8}  & \multicolumn{1}{r|}{49.5} & \multicolumn{1}{r||}{30.3}  & 44.8   \\
\multicolumn{1}{c|}{} & \multicolumn{1}{l|}{\textbf{MUSE}}   & \multicolumn{1}{r|}{47.6}  & \multicolumn{1}{r|}{49.3} & \multicolumn{1}{r|}{48.6}  & \multicolumn{1}{r||}{42.2}  & 46.9   \\
\multicolumn{1}{c|}{} & \multicolumn{1}{l|}{\textbf{Meemi}\textsubscript{VM}}   & \multicolumn{1}{r|}{50.1}  & \multicolumn{1}{r|}{\textbf{53.6}}  & \multicolumn{1}{r|}{\textbf{53.8}}  & \multicolumn{1}{r||}{43.1}  & 50.2   \\
\multicolumn{1}{c|}{} & \multicolumn{1}{l|}{\textbf{Meemi}\textsubscript{MS}}   & \multicolumn{1}{r|}{\textbf{50.4}}  & \multicolumn{1}{r|}{52.5} &  \multicolumn{1}{r|}{52.0}  & \multicolumn{1}{r||}{\textbf{46.6}}  & \textbf{50.4}  \\ \hline

\end{tabular}
}
%}
\caption{Spearman correlation performance of various cross-lingual word embedding models in the cross-lingual word similarity task.}
\label{tab:sim-r}
\end{table}

\subsection{Cross-lingual semantic word similarity}
\label{similarity}

Given a pair of words from two different languages, the task of cross-lingual semantic word similarity consists in measuring to what extent both words are semantically similar. For the evaluation we make use of the cross-lingual word similarity datasets of the SemEval 2017 task \cite{semeval2017similarity}. In this dataset each word from one language is paired with another word from the other language. 
This evaluation task has been found to correlate better with downstream performance than other intrinsic benchmarks \cite{bakarov2018limitations}.
The results are reported in terms of the Pearson and Spearman correlation with respect to human similarity judgments. 
The cross-lingual word similarity results for all the systems are displayed in Table \ref{tab:sim-r}. The languages available for this dataset are English, Spanish, Italian, German and Farsi, hence Finnish and Russian were not evaluated in this task.

\subsection{Cross-lingual natural language inference}
\label{sec:xli}

The task of natural language inference (NLI) consists in detecting entailment, contradiction and neutral relations between pairs of sentences. We test a zero-shot cross-lingual transfer setting where a system is trained with English corpora and is then evaluated on a different language.
It is important to highlight that in this evaluation our main aim is to compare the quality of the cross-lingual word embeddings, and not to develop a state-of-the-art NLI system. Therefore, since this is a downstream task evaluated at the sentence level (and not at the word level as in dictionary induction and semantic word similarity), we develop a simple bag-of-words approach where a sentence embedding is obtained by word vector averaging. 
We then train a linear classifier\footnote{The codebase for these experiments is that of SentEval~\cite{conneau2018senteval}} to obtain the predicted label for each pair of sentences: entailment, contradiction or neutral. We use the full MultiNLI~\cite{williams2018broad} English corpus for training and the Spanish, German and Russian test sets from XNLI~\cite{conneau2018xnli} for testing.
Accuracy results are shown in Table~\ref{tab:xli}.\footnote{For this task we focused on the better performing embeddings learned from Wikipedia and web corpora.}

\begin{table}[!t]
\renewcommand{\arraystretch}{1.02}
%\scalebox{0.75}{
\resizebox{\columnwidth}{!}{
\begin{tabular}{clr|r|r||r}
\hline

\multicolumn{6}{c}{\texttt{\textbf{Wikipedia}}} \\ \hline \hline
\multicolumn{1}{c|}{\textbf{Sup.}} & \multicolumn{1}{c|}{\textbf{Model}} & \multicolumn{1}{c|}{\textbf{EN-ES}} & \multicolumn{1}{c|}{\textbf{EN-DE}} & \multicolumn{1}{c||}{\textbf{EN-RU}} &
\multicolumn{1}{c}{\textbf{Avg}} \\ %\cline{3-5} 
\hline
\multicolumn{1}{c|}{\multirow{2}{*}{\textbf{\texttt{Unsup}}}} & \multicolumn{1}{l|}{\textbf{VecMap}} & \textbf{49.6} & 46.3 & \textbf{34.1} & 43.3\\
\multicolumn{1}{c|}{} & \multicolumn{1}{l|}{\textbf{MUSE}} & 48.4 & 47.4 & 33.3 & 43.0\\
\hline
\multicolumn{1}{c|}{\multirow{2}{*}{\textbf{\texttt{Ident}}}} & \multicolumn{1}{l|}{\textbf{VecMap}} & 43.0 & 42.9 & 33.2 & 39.7\\
\multicolumn{1}{c|}{} & \multicolumn{1}{l|}{\textbf{MUSE}} & 39.5 & 35.8 & 33.3 & 36.2\\
 \hline
\multicolumn{1}{c|}{\multirow{4}{*}{\textbf{\texttt{8K}}}} & \multicolumn{1}{l|}{\textbf{VecMap}} & 49.2 & 46.7 & 33.4 & 43.1 \\
\multicolumn{1}{c|}{} & \multicolumn{1}{l|}{\textbf{MUSE}}  & 47.7 & 47.1 & 33.1 & 42.6\\
\multicolumn{1}{c|}{} & \multicolumn{1}{l|}{\textbf{Meemi}\textsubscript{VM}} & 49.5 & \textbf{47.6} & 33.8 & \textbf{43.6}\\
\multicolumn{1}{c|}{} & \multicolumn{1}{l|}{\textbf{Meemi}\textsubscript{MS}} & 44.2 & 46.7 & 33.3 & 41.4\\ \hline \hline

\multicolumn{6}{c}{\texttt{\textbf{Web corpora}}} \\ \hline \hline
\multicolumn{1}{c|}{\textbf{Sup.}} & \multicolumn{1}{l|}{\textbf{Model}} & \multicolumn{1}{c|}{\textbf{EN-ES}} & \multicolumn{1}{c|}{\textbf{EN-DE}} & \multicolumn{1}{c||}{\textbf{EN-RU}} & 
\multicolumn{1}{c}{\textbf{Avg}} \\ \hline
 
\multicolumn{1}{c|}{\multirow{2}{*}{\textbf{\texttt{Unsup}}}} & \multicolumn{1}{l|}{\textbf{VecMap}} & \textbf{48.5} & 47.9 & 33.4 & 43.3\\
\multicolumn{1}{c|}{} & \multicolumn{1}{l|}{\textbf{MUSE}} & 47.7 & 47.1 & 33.6 & 42.8\\
\hline
\multicolumn{1}{c|}{\multirow{2}{*}{\textbf{\texttt{Ident}}}} & \multicolumn{1}{l|}{\textbf{VecMap}} & 45.5 & 44.4 & 33.4 & 41.1\\
\multicolumn{1}{c|}{} & \multicolumn{1}{l|}{\textbf{MUSE}} & 35.2 & 36.6 & 33.3 & 35.0\\
 \hline
\multicolumn{1}{c|}{\multirow{4}{*}{\textbf{\texttt{8K}}}} & \multicolumn{1}{l|}{\textbf{VecMap}} & 48.4 & 47.5 & 33.2 & 43.0  \\
\multicolumn{1}{c|}{} & \multicolumn{1}{l|}{\textbf{MUSE}}  & 47.3 & \textbf{48.6} & 33.1 & 43.0 \\
\multicolumn{1}{c|}{} & \multicolumn{1}{l|}{\textbf{Meemi}\textsubscript{VM}} & 47.8 & \textbf{48.6} & \textbf{33.8} & \textbf{43.4} \\
\multicolumn{1}{c|}{} & \multicolumn{1}{l|}{\textbf{Meemi}\textsubscript{MS}} & 47.3 & 48.2 & 33.2 & 42.9 \\ \hline \hline

\end{tabular}
}
%}
\caption{Accuracy in the cross-lingual natural language inference task (XNLI) using different cross-lingual word embedding models.}
\label{tab:xli}
\end{table}

\section{Analysis}
\label{analysis}

\begin{figure}
%\centering
\includegraphics[width=\columnwidth]{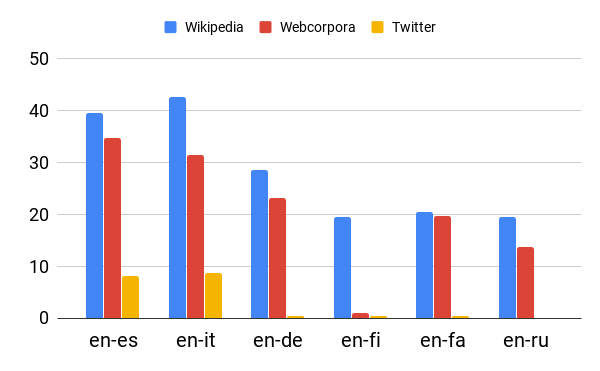}%\includegraphics[width=.525\columnwidth,height=2.25cm]{}\quad
%\medskip
\caption{P@1 performance of the unsupervised version of \texttt{VecMap} on dictionary induction across corpus types and language pairs.}
\label{fig:vecmap_unsup}
\end{figure}

\paragraph{Supervision signals.}

Unsurprisingly, the best alignments of monolingual spaces tend to be obtained with the largest bilingual dictionaries. 
The unsupervised variants of \texttt{VecMap} (see Figure~\ref{fig:vecmap_unsup}) and \texttt{MUSE} attain competitive performance in most cases, especially for comparable corpora where alignments are easier to obtain. However, they struggle in the case of noisy social media corpora and unrelated languages (e.g.\ both \texttt{VecMap} and \texttt{MUSE} obtain inferior results, close to 0, on both Finnish and Farsi), which challenges conclusions from previous work \cite{conneau2018word,artetxe:acl2018}. 
Overall, the results obtained when using social media are clearly inferior, suggesting that there is still room for improvement when it comes to dealing with noisy corpora, regardless of the supervision.

%%%% TABLE SIZE OF DICTIONARY COMPARISON %%%%%
\begin{figure*}[!t]
\hspace{-16px}
%\centering
\includegraphics[width=0.72\columnwidth,height=3.6cm]{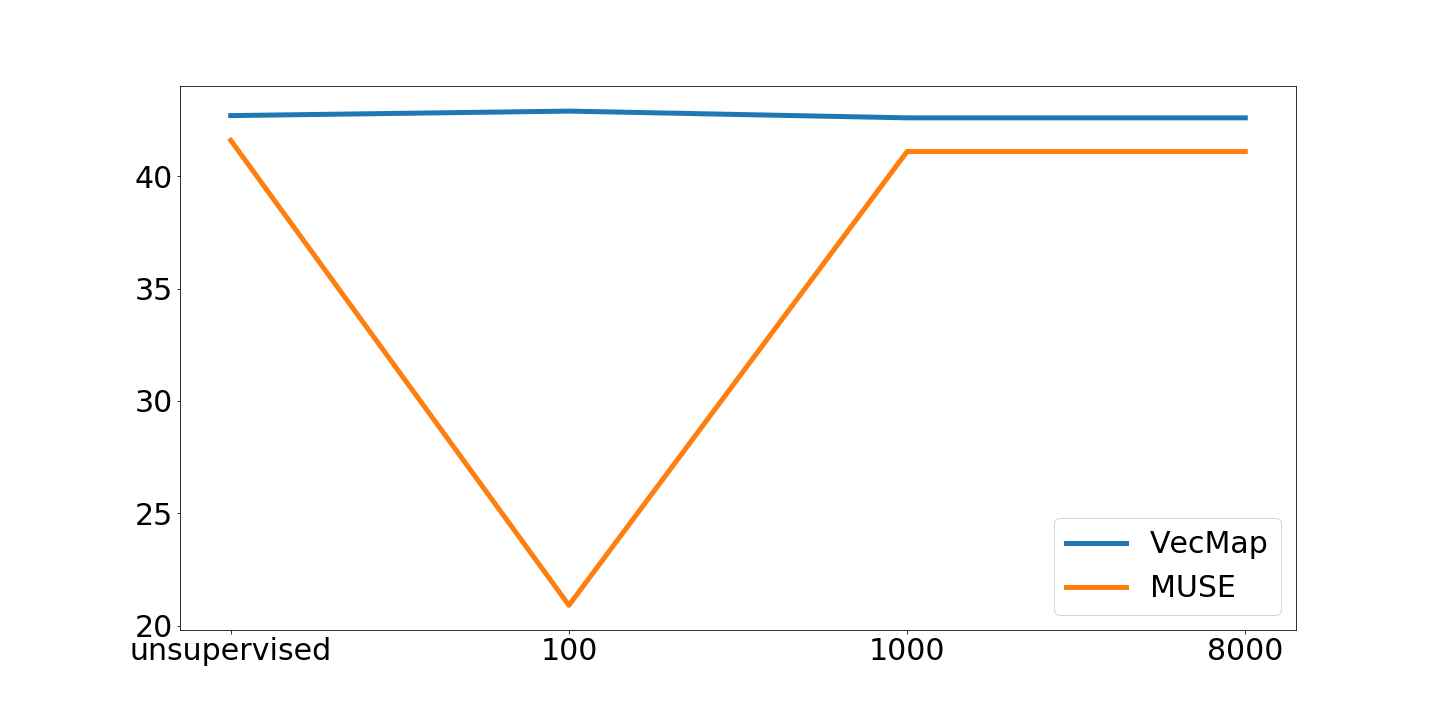}%\includegraphics[width=.525\columnwidth,height=2.25cm]{}\quad
%\medskip
\hspace{0.05cm}
\includegraphics[width=0.72\columnwidth,height=3.6cm]{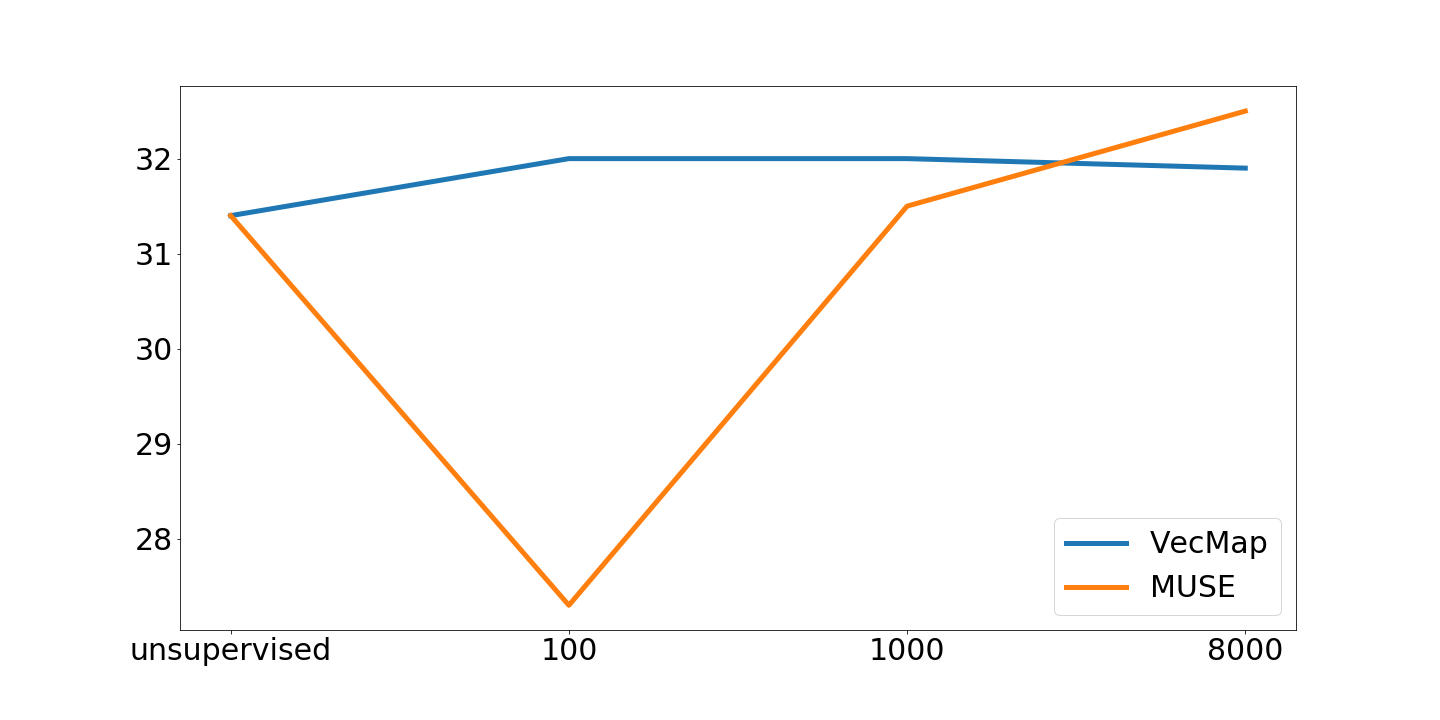}%\includegraphics[width=.525\columnwidth,height=2.25cm]{}\quad
%\medskip
%\hspace{0.3cm}
\includegraphics[width=0.72\columnwidth,height=3.6cm]{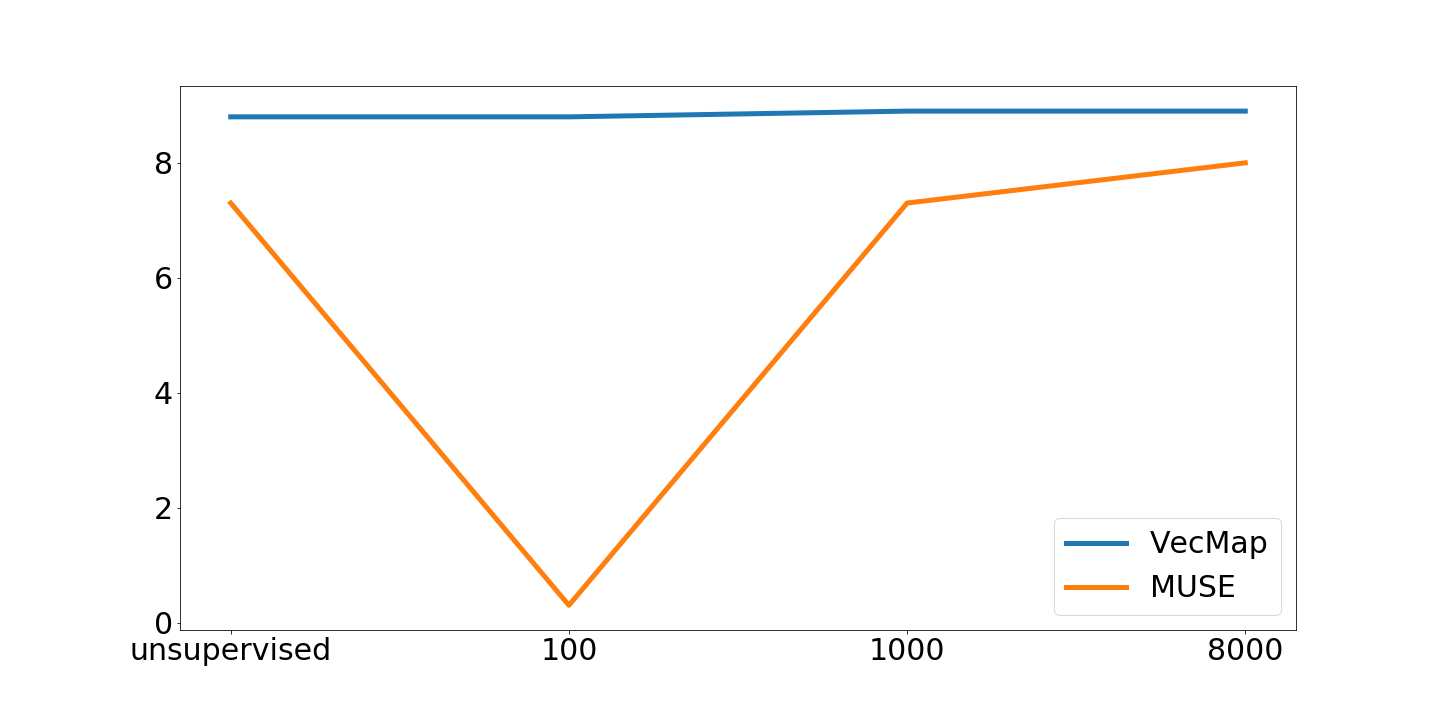}%\includegraphics[width=.525\columnwidth,height=2.25cm]{}\quad
    \caption{Comparison between the dictionary induction performance ($P@1$) of \texttt{VecMap} (blue) and \texttt{MUSE} (red) in English-Italian on Wikipedia (left), web corpora (middle) and social media text (right). The figure clearly shows how \texttt{VecMap} produces similar results irrespective of the seed supervision, while the results of \texttt{MUSE} fluctuate depending on the size of the seed dictionary (with its unsupervised variant being better than using a small dictionary).}
\label{fig:supervision}
\end{figure*}

%%%%%%%%%%%%%%%%%%%%%%%%%%%%%%%%%%%%%%%%%%%%%%

\paragraph{\texttt{VecMap} vs. \texttt{MUSE}.}

One of the main differences between these two models relates to their robustness. The results of \texttt{VecMap} are largely stable across the different types of the supervision. In fact, the best performance for Spanish and Russian on the XNLI task is even obtained in its unsupervised mode. In contrast, \texttt{MUSE} does not perform well with small dictionaries. Figure \ref{fig:supervision} illustrates this trend.
In addition, \texttt{MUSE} also suffers from some stability issues, as it does not always converge to the optimal solution, which confirms findings from previous work \cite{artetxe:acl2018,sogaard2018limitations,hartmann2018unsupervised}.\footnote{This feature was not explicitly tested in this work, as in our experiments models were run until convergence.} In terms of overall results, when given a sufficiently large dictionary training data, the performance of both methods is comparable, which is perhaps unsurprising as they both rely on the solution of the orthogonal Procrustes problem to learn the transformation between the monolingual spaces.

\paragraph{Impact of corpora.}

As can be observed throughout all the experiments, the more comparable and less noisy the monolingual data is, the better the bilingual alignments.
For instance, \texttt{VecMap} goes from an average of $31.2$\% in $P@1$ on Wikipedia down to $4.3$\% on social media, considering all language pairs.
In word similarity, we observe an analogous performance drop, from $68.4$\% to $44.8$\% in Spearman correlation.
Additionally, in Figure~\ref{fig:vecmap_unsup} we can observe the negative influence of noisy corpora and distant languages on the performance of the unsupervised version of \texttt{VecMap} on dictionary induction. 
In terms of error analysis, unsurprisingly we find that the low performance of the models trained on Twitter data is largely due to the noise and the informal nature of the conversation topics. 
For instance, for the word \emph{discover}, instead of \emph{descubren} (one of the correct Spanish translations obtained by the models trained on Wikipedia), the translation given by VecMap corresponds to a misspelling of the correct translation: \emph{descubr}.
As another example, \emph{timeline} is not translated to \emph{cronolog\'ia} in Spanish, but to \emph{instas}, which refers to the social network Instagram. This is clearly due to the specific use of the word \emph{timeline} on Twitter.

\paragraph{Distant languages.}

As expected, the more different the languages are, the harder it is to obtain a reliable alignment of the monolingual spaces. This is particularly noticeable in the case of Farsi, Russian and Finnish (and German to a lesser extent). 
For instance, in the bilingual dictionary task, while most models are over 30.0\% in $P@1$ (excluding social media text which causes performance drops in all languages), for Finnish, Farsi and Russian the results are below 20\% in most cases. A similar tendency can be observed for Farsi on the word similarity task, where the differences are even more pronounced. In addition to its idiosyncrasies (Farsi is considered agglutinative and has a noun compounding formation similar to German), the fact that it uses a different alphabet may explain this large performance gap, noting that FastText takes subword units into account. 
Finally, while the poor performance could be partially explained by the small size of the monolingual training corpora for some languages, it is interesting to see notable performance differences in cases where a distant language has similar or even greater amounts of training data available; e.g., Italian and Russian on Wikipedia, or Italian, Finnish or Russian on web data.

\paragraph{Distant supervision.} As far as the synthetic dictionary of identical words is concerned, \texttt{MUSE} seems to have more difficulties coping with its noisy nature than \texttt{VecMap} (obtaining an average of $16.9$\% versus $23.6$\% in $P@1$ overall in dictionary induction in the Wikipedia and web corpora domains). 
In fact, using \texttt{MUSE} in its unsupervised setting or with a small dictionary generally provides better results. 
However, on social media using the dictionary of identical words appears to help \texttt{MUSE} considerably in the word similarity task compared to the unsupervised setting, going from $19.6$\% to $44.2$\% on average in Spearman correlation overall. This can be attributed to the multilinguality of social media data, where phenomena like code-switching often occur. On the other hand, the consistency of the \texttt{VecMap} semi-supervised algorithm is highlighted again, as using the identical dictionary in this case yields similar results to using external bilingual dictionaries.

\paragraph{Postprocessing.} 

As explained in Section \ref{postprocessing}, for our analysis we experimented with \texttt{Meemi} \cite{doval:meemiemnlp2018}, a recent postprocessing technique which can be applied to any cross-lingual embedding space. There are two main conclusions regarding this technique. First, a clean and relatively big bilingual dictionary is needed in order to get improvements over the base methods \texttt{VecMap} and \texttt{MUSE} (for instance, $+1.2$\% $P@1$ and $+3.1$\% Spearman correlation scores on social media on average, using the 8K dictionary), with the performance otherwise ending up significantly lower. In general, the best overall results are achieved when using this postprocessing technique in combination with the largest dictionary (i.e., 8K pairs). Table \ref{tab:simweb} shows the performance gains or drops by using \texttt{Meemi} in the cross-lingual word similarity task, clearly showing the need for a reasonably large dictionary. This performance variability depending on the size of the dictionary was not addressed in the original paper. Second, \texttt{Meemi} appears to be particularly useful when the monolingual corpora are not comparable, as shown by the larger improvements attained on web-based data. 

\begin{table}[!t]
%\small
\resizebox{\columnwidth}{!}{
\begin{tabular}{|ll|r|r|r|}
\cline{3-5}
\multicolumn{2}{l}{} & \multicolumn{1}{|c|}{\textbf{100}} & \multicolumn{1}{c|}{\textbf{1K}} & \multicolumn{1}{c|}{\textbf{8K}} \\ \hline
\multicolumn{1}{|l|}{\multirow{4}{*}{\textbf{VecMap}}} & \textbf{EN-ES} & -70.4 & -2.0 & \textbf{+0.3} \\ \cline{2-5} 
\multicolumn{1}{|l|}{} & \textbf{EN-IT} & -68.8 & -1.0 & \textbf{+1.2} \\ \cline{2-5}
\multicolumn{1}{|l|}{} & \textbf{EN-DE} & -70.5 & -0.3 & \textbf{+1.4} \\ \cline{2-5} 
\multicolumn{1}{|l|}{} & \textbf{EN-FA} & -30.5 & -32.8 &\textbf{+5.5} \\ \hline
\multicolumn{1}{|l|}{\multirow{4}{*}{\textbf{MUSE}}} & \textbf{EN-ES} & -71.5 & -1.8 & \textbf{+1.7} \\ \cline{2-5} 
\multicolumn{1}{|l|}{} & \textbf{EN-IT} & -65.1 & -0.8 & \textbf{+0.7} \\ \cline{2-5} 
\multicolumn{1}{|l|}{} & \textbf{EN-DE} & -68.8 & -0.1 & \textbf{+1.9} \\ \cline{2-5} 
\multicolumn{1}{|l|}{} & \textbf{EN-FA} & -7.4 & -22.3 & \textbf{+9.1} \\ \hline
\end{tabular}
}
\caption{Absolute improvement (in percentage points) by applying the postprocessing (\texttt{Meemi}) over the two base models VecMap and MUSE on the cross-lingual word similarity task using web corpora.}
\label{tab:simweb}
\end{table}

\paragraph{Evaluation tasks.}
The performance variability in bilingual dictionary induction, cross-lingual word similarity and cross-lingual inference seems to be very similar across the board, with the main difference being the lower variability in results in the cross-lingual NLI task (which can be expected given that it is a downstream task where additional factors are also involved). The factors with the greatest impact on performance, namely monolingual corpora and language pairs, are clearly reflected in both cases, with analogous drops when going from training on Wikipedia to social media, and also when testing on Finnish, Farsi or Russian. To test our intuition, we computed Pearson correlation values from all overlapping results between task pairs. In this case, similarity and dictionary induction attain the highest correlation ($r=0.78$), with cross-lingual NLI and dictionary induction also attaining a high correlation score ($r=0.73$). The lowest correlation score corresponds to cross-lingual similarity and NLI, with a lower figure of $r=0.28$. Despite being positive, this relatively low correlation may suggest that dictionary induction would be a better proxy to test cross-lingual embedding performance in downstream tasks. We should note, however, that these correlation figures are only indicative and particular to the methods tested in our analysis and therefore should not be taken as the global correlation between tasks.

\section{Conclusions}

We have presented an extensive evaluation of state-of-the-art cross-lingual embedding models in a wide variety of experimental settings. 
The variables explored in this paper were: the choice of training corpus, the type of  supervision signal (including different types of bilingual dictionaries), and the language pairs considered. Likewise, the evaluation procedure included two standard benchmarks for cross-lingual embedding evaluation, namely bilingual dictionary induction and cross-lingual word similarity, as well as cross-lingual natural language inference as an extrinsic task. The set of languages considered included not only related languages such as English, Spanish, Italian and German, but also languages from different families such as Finnish, Farsi and Russian.

Our analysis highlights a particularly marked variability in the performance of the considered methods concerning (1) the monolingual training corpora used (e.g., between comparable corpora such as Wikipedia and non-comparable or noisy user-generated corpora) and (2) language pairs (distant language pairs still constitute a major challenge). We may also conclude that bilingual supervision signals constitute a key component for most models in non-ideal settings (i.e., non-comparable corpora or distant languages). 
In general, our analysis and the results show that supervised cross-lingual word embedding learning is more robust than purely unsupervised cross-lingual learning, challenging claims from previous works on this regard \cite{conneau2018word,artetxe:acl2018,chen2018unsupervised,hoshen-wolf-2018-non,xu2018crosslingual} and in line with a concurring analysis showing a similar trend \cite{vulic2019we}.

As future work, it would be interesting to analyze multilingual embeddings that involve more than two languages, along the lines of recent multilingual approaches \cite{chen2018unsupervised,heyman-etal-2019-learning,doval2019meemi}.

\section{Acknowledgments}

Steven Schockaert is supported by ERC Starting Grant 637277. Yerai Doval has been supported by the Spanish Ministry of Economy, Industry and Competitiveness (MINECO) through the ANSWER-ASAP project (TIN2017-85160-C2-2-R); by the Spanish State Secretariat for Research, Development and Innovation (which belongs to MINECO) and the European Social Fund (ESF) through a FPI fellowship (BES-2015-073768) associated to TELEPARES project (FFI2014-51978-C2-1-R); and by the Xunta de Galicia through TELGALICIA research network (ED431D 2017/12).

\section*{Appendix: Additional Experimental Results}
\label{sec:appendix}

This appendix contains additional experimental results not included in the main body of the paper. In particular, it contains supplementary results for the dictionary induction (Table \ref{tab:dictionary}) and cross-lingual word similarity (Table \ref{tab:sim}) tasks, using all sources of supervision: no supervision, dictionary of identical words, and dictionaries containing 100, 1K and 8K translation pairs. The methods included in these tables are explained in the paper (Section \ref{methods}).

\begin{table*}[!t]
\renewcommand{\arraystretch}{1.2}
\resizebox{\textwidth}{!}{
\begin{tabular}{llrrrrrrrrrrrrrrr}
\hline
\multicolumn{17}{c}{\textbf{\texttt{Wikipedia}}} \\ \hline
\multicolumn{1}{l|}{\multirow{2}{*}{\textbf{Model}}} & \multicolumn{1}{l|}{\multirow{2}{*}{\textbf{Supervision}}} & \multicolumn{3}{c}{\textbf{English-Spanish}} & \multicolumn{3}{c|}{\textbf{English-Italian}} & \multicolumn{3}{c|}{\textbf{English-German}} & \multicolumn{3}{c|}{\textbf{English-Finnish}} & \multicolumn{3}{c}{\textbf{English-Farsi}} \\ \cline{3-17} 
\multicolumn{1}{l|}{} & \multicolumn{1}{l|}{} & \multicolumn{1}{l}{$P@1$} & \multicolumn{1}{l}{$P@5$} & \multicolumn{1}{l|}{$P@10$} & \multicolumn{1}{l}{$P@1$} & \multicolumn{1}{l}{$P@5$} & \multicolumn{1}{l|}{$P@10$} & \multicolumn{1}{l}{$P@1$} & \multicolumn{1}{l}{$P@5$} & \multicolumn{1}{l|}{$P@10$} & \multicolumn{1}{l}{$P@1$} & \multicolumn{1}{l}{$P@5$} & \multicolumn{1}{l|}{$P@10$} & \multicolumn{1}{l}{$P@1$} & \multicolumn{1}{l}{$P@5$} & \multicolumn{1}{l}{$P@10$} \\ \hline
\multicolumn{1}{l|}{\multirow{5}{*}{\textbf{\texttt{VecMap}}}} & \multicolumn{1}{l|}{\textbf{8K}} & \textbf{39.6} & 66.2 & \multicolumn{1}{r|}{72.3} & 42.6 & 65.9 & \multicolumn{1}{r|}{71.8} & 28.6 & \textbf{48.3} & \multicolumn{1}{r|}{\textbf{54.8}} & 22.4 & 44.5 & \multicolumn{1}{r|}{52.5} & 22.8 & 39.7 & 46.2 \\
\multicolumn{1}{l|}{} & \multicolumn{1}{l|}{\textbf{1K}} & \textbf{39.6} & 66.2 & \multicolumn{1}{r|}{72.3} & 42.6 & 65.7 & \multicolumn{1}{r|}{71.6} & \textbf{28.7} & \textbf{48.3} & \multicolumn{1}{r|}{54.7} & 22.2 & 43.9 & \multicolumn{1}{r|}{51.7} & 23.2 & 40.2 & 46.1 \\
\multicolumn{1}{l|}{} & \multicolumn{1}{l|}{\textbf{100}} & \textbf{39.6} & 66.2 & \multicolumn{1}{r|}{72.4} & \textbf{42.9} & 65.7 & \multicolumn{1}{r|}{71.6} & 28.6 & \textbf{48.3} & \multicolumn{1}{r|}{\textbf{54.8}} & 21.6 & 43.4 & \multicolumn{1}{r|}{51.7} & 22.7 & \textbf{40.6} & \textbf{46.4} \\
\multicolumn{1}{l|}{} & \multicolumn{1}{l|}{\textbf{Identical}} & 39.5 & 66.0 & \multicolumn{1}{r|}{72.4} & 42.7 & 65.8 & \multicolumn{1}{r|}{71.7} & 28.6 & \textbf{48.3} & \multicolumn{1}{r|}{54.7} & 21.6 & 43.7 & \multicolumn{1}{r|}{51.6} & \textbf{23.4} & 40.3 & 46.1 \\
\multicolumn{1}{l|}{} & \multicolumn{1}{l|}{\textbf{Unsupervised}} & \textbf{39.6} & 66.1 & \multicolumn{1}{r|}{72.3} & 42.7 & 65.7 & \multicolumn{1}{r|}{71.6} & 28.6 & \textbf{48.3} & \multicolumn{1}{r|}{\textbf{54.8}} & 19.6 & 40.4 & \multicolumn{1}{r|}{48.3} & 20.5 & 37.0 & 42.8 \\ \hline

\multicolumn{1}{l|}{\multirow{5}{*}{\textbf{\texttt{MUSE}}}} & \multicolumn{1}{l|}{\textbf{8K}} & 39.1 & 65.4 & \multicolumn{1}{r|}{72.3} & 41.1 & 63.3 & \multicolumn{1}{r|}{70.1} & 27.6 & 45.9 & \multicolumn{1}{r|}{53.2} & 19.5 & 40.4 & \multicolumn{1}{r|}{49.5} & 19.7 & 35.4 & 42 \\
\multicolumn{1}{l|}{} & \multicolumn{1}{l|}{\textbf{1K}} & 39.2 & 65.4 & \multicolumn{1}{r|}{72.1} & 41.1 & 63.3 & \multicolumn{1}{r|}{70.1} & 27.6 & 46.0 & \multicolumn{1}{r|}{53.1} & 18.1 & 36.8 & \multicolumn{1}{r|}{44.9} & 19.8 & 35.3 & 41.5 \\
\multicolumn{1}{l|}{} & \multicolumn{1}{l|}{\textbf{100}} & 24.8 & 47.5 & \multicolumn{1}{r|}{54.6} & 20.9 & 39.2 & \multicolumn{1}{r|}{48.1} & 0.8 & 3.4 & \multicolumn{1}{r|}{5.2} & 0.3 & 1.3 & \multicolumn{1}{r|}{2.2} & 6.2 & 16.1 & 22.8 \\
\multicolumn{1}{l|}{} & \multicolumn{1}{l|}{\textbf{Identical}} & 35.9 & 60.6 & \multicolumn{1}{r|}{67.3} & 37.8 & 60.4 & \multicolumn{1}{r|}{68.5} & 24.8 & 41.9 & \multicolumn{1}{r|}{49.5} & 13.4 & 25.5 & \multicolumn{1}{r|}{32} & 6.7 & 16.6 & 21.3 \\
\multicolumn{1}{l|}{} & \multicolumn{1}{l|}{\textbf{Unsupervised}} & 39.3 & 64.7 & \multicolumn{1}{r|}{71.3} & 41.6 & 63.2 & \multicolumn{1}{r|}{69.9} & 28.3 & 46.5 & \multicolumn{1}{r|}{53.3} & 0.0 & 0.0 & \multicolumn{1}{r|}{0.0} & 0.0 & 0.0 & 0.0 \\ \hline

\multicolumn{1}{l|}{\multirow{4}{*}{\textbf{\textbf{Meemi}\textsubscript{VM}}}} & \multicolumn{1}{l|}{\textbf{8K}} & 39.3 & \textbf{67.4} & \multicolumn{1}{r|}{\textbf{73.7}} & 41.6 & 66.5 & \multicolumn{1}{r|}{72.5} & 28 & 47.8 & \multicolumn{1}{r|}{\textbf{54.8}} & \textbf{23.8} & \textbf{48.7} & \multicolumn{1}{r|}{\textbf{57.0}} & 0.0 & 0.0 & 0.0 \\
\multicolumn{1}{l|}{} & \multicolumn{1}{l|}{\textbf{1K}} & 35.5 & 63.7 & \multicolumn{1}{r|}{69.4} & 38.6 & 64.0 & \multicolumn{1}{r|}{70.1} & 23.1 & 42.5 & \multicolumn{1}{r|}{49.9} & 17.8 & 40.1 & \multicolumn{1}{r|}{48.6} & 0.0 & 0.0 & 0.0 \\
\multicolumn{1}{l|}{} & \multicolumn{1}{l|}{\textbf{100}} & 0.0 & 0.0 & \multicolumn{1}{r|}{0.0} & 0.0 & 0.0 & \multicolumn{1}{r|}{0.0} & 0.0 & 0.0 & \multicolumn{1}{r|}{0.0} & 0.0 & 0.0 & \multicolumn{1}{r|}{0.0} & 0.0 & 0.0 & 0.0 \\
\multicolumn{1}{l|}{} & \multicolumn{1}{l|}{\textbf{Identical}} & 38.7 & 63.7 & \multicolumn{1}{r|}{70.1} & 40.6 & 64.1 & \multicolumn{1}{r|}{70.2} & 27.5 & 46.6 & \multicolumn{1}{r|}{53.1} & 19.3 & 37.7 & \multicolumn{1}{r|}{45.5} & 7.1 & 14.3 & 17.9 \\ \hline
\multicolumn{1}{l|}{\multirow{4}{*}{\textbf{\textbf{Meemi}\textsubscript{MS}}}} & \multicolumn{1}{l|}{\textbf{8K}} & 39.3 & \textbf{67.4} & \multicolumn{1}{r|}{\textbf{73.7}} & 41.3 & \textbf{66.8} & \multicolumn{1}{r|}{\textbf{72.8}} & 27.1 & 46.3 & \multicolumn{1}{r|}{53.9} & 21.7 & 45.0 & \multicolumn{1}{r|}{53.6} & 0.0 & 0.0 & 0.0 \\
\multicolumn{1}{l|}{} & \multicolumn{1}{l|}{\textbf{1K}} & 35.4 & 63.1 & \multicolumn{1}{r|}{69.3} & 38.2 & 63.6 & \multicolumn{1}{r|}{70.2} & 22.4 & 40.4 & \multicolumn{1}{r|}{47.9} & 14.7 & 33.6 & \multicolumn{1}{r|}{41.8} & 0.0 & 0.0 & 0.0 \\
\multicolumn{1}{l|}{} & \multicolumn{1}{l|}{\textbf{100}} & 0.0 & 0.0 & \multicolumn{1}{r|}{0.0} & 0.0 & 0.1 & \multicolumn{1}{r|}{0.1} & 0.0 & 0.0 & \multicolumn{1}{r|}{0.0} & 0.0 & 0.0 & \multicolumn{1}{r|}{0.0} & 0.0 & 0.0 & 0.0 \\
\multicolumn{1}{l|}{} & \multicolumn{1}{l|}{\textbf{Identical}} & 35.4 & 58.9 & \multicolumn{1}{r|}{65.4} & 37.0 & 59.0 & \multicolumn{1}{r|}{65.9} & 24.0 & 40.0 & \multicolumn{1}{r|}{47.0} & 13.0 & 25.5 & \multicolumn{1}{r|}{32.0} & 2.5 & 6.2 & 8.3 \\ \hline

\multicolumn{17}{c}{\textbf{\texttt{Web corpora}}} \\ \hline

\multicolumn{1}{l|}{\multirow{2}{*}{\textbf{Model}}} & \multicolumn{1}{l|}{\multirow{2}{*}{\textbf{Supervision}}} & \multicolumn{3}{c}{\textbf{English-Spanish}} & \multicolumn{3}{c|}{\textbf{English-Italian}} & \multicolumn{3}{c|}{\textbf{English-German}} & \multicolumn{3}{c|}{\textbf{English-Finnish}} & \multicolumn{3}{c}{\textbf{English-Farsi}} \\ \cline{3-17} 
\multicolumn{1}{l|}{} & \multicolumn{1}{l|}{} & \multicolumn{1}{l}{$P@1$} & \multicolumn{1}{l}{$P@5$} & \multicolumn{1}{l|}{$P@10$} & \multicolumn{1}{l}{$P@1$} & \multicolumn{1}{l}{$P@5$} & \multicolumn{1}{l|}{$P@10$} & \multicolumn{1}{l}{$P@1$} & \multicolumn{1}{l}{$P@5$} & \multicolumn{1}{l|}{$P@10$} & \multicolumn{1}{l}{$P@1$} & \multicolumn{1}{l}{$P@5$} & \multicolumn{1}{l|}{$P@10$} & \multicolumn{1}{l}{$P@1$} & \multicolumn{1}{l}{$P@5$} & \multicolumn{1}{l}{$P@10$} \\ \hline
\multicolumn{1}{l|}{\multirow{5}{*}{\textbf{\texttt{VecMap}}}} & \multicolumn{1}{l|}{\textbf{8K}} & 34.6 & 60.6 & \multicolumn{1}{r|}{66.9} & 31.9 & 54.2 & \multicolumn{1}{r|}{60.4} & 23.1 & 42.7 & \multicolumn{1}{r|}{50.5} & 18.9 & 40.9 & \multicolumn{1}{r|}{48.8} & 19.6 & \textbf{35.8} & \textbf{41.4} \\
\multicolumn{1}{l|}{} & \multicolumn{1}{l|}{\textbf{1K}} & 34.6 & 60.5 & \multicolumn{1}{r|}{67.0} & 32.0 & 54.0 & \multicolumn{1}{r|}{60.5} & 23.1 & 42.7 & \multicolumn{1}{r|}{50.3} & 19.4 & 42.0 & \multicolumn{1}{r|}{49.4} & 19.5 & 35.6 & 41.2 \\
\multicolumn{1}{l|}{} & \multicolumn{1}{l|}{\textbf{100}} & \textbf{38.5} & 61.2 & \multicolumn{1}{r|}{67.5} & 32.0 & 54.2 & \multicolumn{1}{r|}{60.5} & 23.0 & 43.0 & \multicolumn{1}{r|}{50.2} & 19.3 & 41.6 & \multicolumn{1}{r|}{49.6} & 19.7 & 35.5 & 41.3 \\
\multicolumn{1}{l|}{} & \multicolumn{1}{l|}{\textbf{Identical}} & 34.7 & 60.4 & \multicolumn{1}{r|}{67.0} & 31.4 & 54.0 & \multicolumn{1}{r|}{60.7} & 23.1 & 42.9 & \multicolumn{1}{r|}{50.5} & 18.6 & 41.6 & \multicolumn{1}{r|}{49.3} & \textbf{20.0} & 35.3 & 40.3 \\
\multicolumn{1}{l|}{} & \multicolumn{1}{l|}{\textbf{Unsupervised}} & 34.8 & 60.6 & \multicolumn{1}{r|}{67.0} & 31.4 & 53.7 & \multicolumn{1}{r|}{60.7} & 23.2 & 42.7 & \multicolumn{1}{r|}{50.2} & 0.0 & 0.0 & \multicolumn{1}{r|}{0.0} & 19.7 & 34.6 & 40.4 \\ \hline
\multicolumn{1}{l|}{\multirow{5}{*}{\textbf{\texttt{MUSE}}}} & \multicolumn{1}{l|}{\textbf{8K}} & 32.5 & 58.2 & \multicolumn{1}{r|}{65.9} & 32.5 & 56.0 & \multicolumn{1}{r|}{63.2} & 22.4 & 40.9 & \multicolumn{1}{r|}{48.9} & 20.0 & 40.1 & \multicolumn{1}{r|}{48.3} & 17.4 & 31.6 & 37.6 \\
\multicolumn{1}{l|}{} & \multicolumn{1}{l|}{\textbf{1K}} & 32.9 & 56.8 & \multicolumn{1}{r|}{64.2} & 31.5 & 52.7 & \multicolumn{1}{r|}{60.6} & 22.1 & 41.0 & \multicolumn{1}{r|}{48.2} & 18.4 & 39.1 & \multicolumn{1}{r|}{47.7} & 16.6 & 31.1 & 36.4 \\
\multicolumn{1}{l|}{} & \multicolumn{1}{l|}{\textbf{100}} & 32.3 & 56.1 & \multicolumn{1}{r|}{63.9} & 27.3 & 48.0 & \multicolumn{1}{r|}{55.3} & 17.8 & 35.0 & \multicolumn{1}{r|}{41.6} & 2.7 & 7.9 & \multicolumn{1}{r|}{11.1} & 0.0 & 0.5 & 0.7 \\
\multicolumn{1}{l|}{} & \multicolumn{1}{l|}{\textbf{Identical}} & 26.1 & 46.7 & \multicolumn{1}{r|}{53.8} & 24.7 & 45.1 & \multicolumn{1}{r|}{52.4} & 17.4 & 32.8 & \multicolumn{1}{r|}{40.5} & 12.6 & 26.0 & \multicolumn{1}{r|}{33.8} & 3.0 & 8.3 & 5.8 \\
\multicolumn{1}{l|}{} & \multicolumn{1}{l|}{\textbf{Unsupervised}} & 31.4 & 51.2 & \multicolumn{1}{r|}{57.7} & 31.4 & 51.2 & \multicolumn{1}{r|}{57.7} & 20.8 & 38.7 & \multicolumn{1}{r|}{46.6} & 0.0 & 0.0 & \multicolumn{1}{r|}{0.0} & 18.1 & 32.8 & 37.8 \\ \hline
\multicolumn{1}{l|}{\multirow{4}{*}{\textbf{\begin{tabular}[c]{@{}l@{}}\textbf{Meemi}\textsubscript{VM}\end{tabular}}}} & \multicolumn{1}{l|}{\textbf{8K}} & 34.5 & \textbf{61.6} & \multicolumn{1}{r|}{67.9} & 33.6 & 58.3 & \multicolumn{1}{r|}{\textbf{65.6}} & \textbf{23.7} & \textbf{45.4} & \multicolumn{1}{r|}{\textbf{53.2}} & 22.3 & \textbf{46.7} & \multicolumn{1}{r|}{\textbf{55.0}} & 0.0 & 0.0 & 0.0 \\
\multicolumn{1}{l|}{} & \multicolumn{1}{l|}{\textbf{1K}} & 30.2 & 55.0 & \multicolumn{1}{r|}{62.7} & 30.7 & 54.0 & \multicolumn{1}{r|}{61.1} & 19.4 & 38.9 & \multicolumn{1}{r|}{45.9} & 18.2 & 39.9 & \multicolumn{1}{r|}{48.1} & 0.0 & 0.0 & 0.0 \\
\multicolumn{1}{l|}{} & \multicolumn{1}{l|}{\textbf{100}} & 0.0 & 0.0 & \multicolumn{1}{r|}{0.0} & 0.0 & 0.0 & \multicolumn{1}{r|}{0.0} & 0.0 & 0.0 & \multicolumn{1}{r|}{0.0} & 0.0 & 0.0 & \multicolumn{1}{r|}{0.0} & 0.0 & 0.0 & 0.0 \\
\multicolumn{1}{l|}{} & \multicolumn{1}{l|}{\textbf{Identical}} & 34.1 & 58.3 & \multicolumn{1}{r|}{64.8} & 31.6 & 54.6 & \multicolumn{1}{r|}{62.5} & 22.5 & 42.0 & \multicolumn{1}{r|}{49.0} & 21.1 & 43.2 & \multicolumn{1}{r|}{51.3} & 11.2 & 23.9 & 28.6 \\ \hline
\multicolumn{1}{l|}{\multirow{4}{*}{\textbf{\begin{tabular}[c]{@{}l@{}}\textbf{Meemi}\textsubscript{MS}\end{tabular}}}} & \multicolumn{1}{l|}{\textbf{8K}} & 33.9 & 60.7 & \multicolumn{1}{r|}{\textbf{68.4}} & \textbf{33.8} & \textbf{58.4} & \multicolumn{1}{r|}{\textbf{65.6}} & \textbf{23.7} & 45.3 & \multicolumn{1}{r|}{52.3} & \textbf{23.0} & 46.1 & \multicolumn{1}{r|}{54.0} & 0.0 & 0.0 & 0.0 \\
\multicolumn{1}{l|}{} & \multicolumn{1}{l|}{\textbf{1K}} & 29.1 & 54.6 & \multicolumn{1}{r|}{62.3} & 29.9 & 52.7 & \multicolumn{1}{r|}{60.3} & 18.3 & 37.0 & \multicolumn{1}{r|}{44.1} & 17.2 & 37.4 & \multicolumn{1}{r|}{45.6} & 0.0 & 0.0 & 0.0 \\
\multicolumn{1}{l|}{} & \multicolumn{1}{l|}{\textbf{100}} & 0.0 & 0.0 & \multicolumn{1}{r|}{0.0} & 0.0 & 0.0 & \multicolumn{1}{r|}{0.0} & 0.0 & 0.0 & \multicolumn{1}{r|}{0.0} & 0.0 & 0.0 & \multicolumn{1}{r|}{0.0} & 0.0 & 0.0 & 0.0 \\
\multicolumn{1}{l|}{} & \multicolumn{1}{l|}{\textbf{Identical}} & 24.3 & 43.0 & \multicolumn{1}{r|}{49.8} & 21.8 & 41.1 & \multicolumn{1}{r|}{48.9} & 16.4 & 30.7 & \multicolumn{1}{r|}{37.3} & 13.1 & 26.1 & \multicolumn{1}{r|}{33.9} & 2.0 & 4.2 & 5.8 \\ \hline

\multicolumn{17}{c}{\textbf{\texttt{Social media}}} \\ \hline \hline

\multicolumn{1}{l|}{\multirow{2}{*}{\textbf{Model}}} & \multicolumn{1}{l|}{\multirow{2}{*}{\textbf{Supervision}}} & \multicolumn{3}{c}{\textbf{English-Spanish}} & \multicolumn{3}{c|}{\textbf{English-Italian}} & \multicolumn{3}{c|}{\textbf{English-German}} & \multicolumn{3}{c|}{\textbf{English-Finnish}} & \multicolumn{3}{c|}{\textbf{English-Farsi}} \\ \cline{3-17} 
\multicolumn{1}{l|}{} & \multicolumn{1}{l|}{} & $P@1$ & $P@5$ & \multicolumn{1}{l|}{$P@10$} & $P@1$ & $P@5$ & \multicolumn{1}{l|}{$P@10$} & $P@1$ & $P@5$ & \multicolumn{1}{l|}{$P@10$} & $P@1$ & $P@5$ & \multicolumn{1}{l|}{$P@10$} & $P@1$ & $P@5$ & $P@10$ \\ \hline
\multicolumn{1}{l|}{\multirow{5}{*}{\textbf{\texttt{VecMap}}}} & \multicolumn{1}{l|}{\textbf{8K}} & \multicolumn{1}{r}{8.7} & \multicolumn{1}{r}{16.6} & \multicolumn{1}{r|}{21.6} & \multicolumn{1}{r}{8.9} & \multicolumn{1}{r}{17.3} & \multicolumn{1}{r|}{22.4} & \multicolumn{1}{r}{3.2} & \multicolumn{1}{r}{6.8} & \multicolumn{1}{r|}{9.5} & \multicolumn{1}{r}{0.2} & \multicolumn{1}{r}{0.8} & \multicolumn{1}{r|}{1.2} & \multicolumn{1}{r}{0.4} & \multicolumn{1}{r}{1.6} & \multicolumn{1}{r}{2.0} \\
\multicolumn{1}{l|}{} & \multicolumn{1}{l|}{\textbf{1K}} & \multicolumn{1}{r}{8.3} & \multicolumn{1}{r}{17.0} & \multicolumn{1}{r|}{21.3} & \multicolumn{1}{r}{8.9} & \multicolumn{1}{r}{17.5} & \multicolumn{1}{r|}{22.0} & \multicolumn{1}{r}{2.9} & \multicolumn{1}{r}{6.5} & \multicolumn{1}{r|}{9.3} & \multicolumn{1}{r}{0.0} & \multicolumn{1}{r}{0.4} & \multicolumn{1}{r|}{1.1} & \multicolumn{1}{r}{0.3} & \multicolumn{1}{r}{1.0} & \multicolumn{1}{r}{1.4} \\
\multicolumn{1}{l|}{} & \multicolumn{1}{l|}{\textbf{100}} & \multicolumn{1}{r}{7.9} & \multicolumn{1}{r}{15.9} & \multicolumn{1}{r|}{20.2} & \multicolumn{1}{r}{8.8} & \multicolumn{1}{r}{17.6} & \multicolumn{1}{r|}{22.3} & \multicolumn{1}{r}{2.8} & \multicolumn{1}{r}{6.0} & \multicolumn{1}{r|}{8.6} & \multicolumn{1}{r}{0.0} & \multicolumn{1}{r}{0.0} & \multicolumn{1}{r|}{0.1} & \multicolumn{1}{r}{0.1} & \multicolumn{1}{r}{0.3} & \multicolumn{1}{r}{0.4} \\
\multicolumn{1}{l|}{} & \multicolumn{1}{l|}{\textbf{Identical}} & \multicolumn{1}{r}{8.5} & \multicolumn{1}{r}{16.9} & \multicolumn{1}{r|}{21.6} & \multicolumn{1}{r}{9.1} & \multicolumn{1}{r}{16.8} & \multicolumn{1}{r|}{21.8} & \multicolumn{1}{r}{2.6} & \multicolumn{1}{r}{6.7} & \multicolumn{1}{r|}{9.6} & \multicolumn{1}{r}{0.0} & \multicolumn{1}{r}{0.0} & \multicolumn{1}{r|}{0.0} & \multicolumn{1}{r}{0.2} & \multicolumn{1}{r}{0.5} & \multicolumn{1}{r}{1.1} \\
\multicolumn{1}{l|}{} & \multicolumn{1}{l|}{\textbf{Unsupervised}} & \multicolumn{1}{r}{8.1} & \multicolumn{1}{r}{16.4} & \multicolumn{1}{r|}{20.4} & \multicolumn{1}{r}{8.8} & \multicolumn{1}{r}{17.0} & \multicolumn{1}{r|}{22.3} & \multicolumn{1}{r}{0.1} & \multicolumn{1}{r}{0.4} & \multicolumn{1}{r|}{0.5} & \multicolumn{1}{r}{0.0} & \multicolumn{1}{r}{0.0} & \multicolumn{1}{r|}{0.0} & \multicolumn{1}{r}{0.0} & \multicolumn{1}{r}{0.0} & \multicolumn{1}{r}{0.0} \\ \hline
\multicolumn{1}{l|}{\multirow{5}{*}{\textbf{\texttt{MUSE}}}} & \multicolumn{1}{l|}{\textbf{8K}} & \multicolumn{1}{r}{8.1} & \multicolumn{1}{r}{17.6} & \multicolumn{1}{r|}{22.7} & \multicolumn{1}{r}{8.0} & \multicolumn{1}{r}{16.4} & \multicolumn{1}{r|}{21.1} & \multicolumn{1}{r}{2.2} & \multicolumn{1}{r}{6.0} & \multicolumn{1}{r|}{8.4} & \multicolumn{1}{r}{0.6} & \multicolumn{1}{r}{2.2} & \multicolumn{1}{r|}{3.2} & \multicolumn{1}{r}{\textbf{1.2}} & \multicolumn{1}{r}{\textbf{4.5}} & \multicolumn{1}{r}{\textbf{6.3}} \\
\multicolumn{1}{l|}{} & \multicolumn{1}{l|}{\textbf{1K}} & \multicolumn{1}{r}{7.2} & \multicolumn{1}{r}{15.9} & \multicolumn{1}{r|}{20.5} & \multicolumn{1}{r}{7.3} & \multicolumn{1}{r}{14.6} & \multicolumn{1}{r|}{18.4} & \multicolumn{1}{r}{0.9} & \multicolumn{1}{r}{3.0} & \multicolumn{1}{r|}{4.5} & \multicolumn{1}{r}{0.6} & \multicolumn{1}{r}{1.5} & \multicolumn{1}{r|}{2.1} & \multicolumn{1}{r}{0.9} & \multicolumn{1}{r}{2.1} & \multicolumn{1}{r}{3.4} \\
\multicolumn{1}{l|}{} & \multicolumn{1}{l|}{\textbf{100}} & \multicolumn{1}{r}{0.4} & \multicolumn{1}{r}{1.1} & \multicolumn{1}{r|}{1.9} & \multicolumn{1}{r}{0.3} & \multicolumn{1}{r}{1.1} & \multicolumn{1}{r|}{1.8} & \multicolumn{1}{r}{0.1} & \multicolumn{1}{r}{0.3} & \multicolumn{1}{r|}{0.6} & \multicolumn{1}{r}{0.0} & \multicolumn{1}{r}{0.2} & \multicolumn{1}{r|}{0.4} & \multicolumn{1}{r}{0.1} & \multicolumn{1}{r}{0.3} & \multicolumn{1}{r}{0.4} \\
\multicolumn{1}{l|}{} & \multicolumn{1}{l|}{\textbf{Identical}} & \multicolumn{1}{r}{2.5} & \multicolumn{1}{r}{5.2} & \multicolumn{1}{r|}{7.1} & \multicolumn{1}{r}{3.9} & \multicolumn{1}{r}{10.1} & \multicolumn{1}{r|}{13.7} & \multicolumn{1}{r}{1.1} & \multicolumn{1}{r}{2.6} & \multicolumn{1}{r|}{3.7} & \multicolumn{1}{r}{0.1} & \multicolumn{1}{r}{0.1} & \multicolumn{1}{r|}{0.2} & \multicolumn{1}{r}{0.1} & \multicolumn{1}{r}{0.3} & \multicolumn{1}{r}{0.8} \\
\multicolumn{1}{l|}{} & \multicolumn{1}{l|}{\textbf{Unsupervised}} & \multicolumn{1}{r}{0.0} & \multicolumn{1}{r}{0.0} & \multicolumn{1}{r|}{0.0} & \multicolumn{1}{r}{7.3} & \multicolumn{1}{r}{14.5} & \multicolumn{1}{r|}{18.3} & \multicolumn{1}{r}{0.0} & \multicolumn{1}{r}{0.0} & \multicolumn{1}{r|}{0.0} & \multicolumn{1}{r}{0.0} & \multicolumn{1}{r}{0.0} & \multicolumn{1}{r|}{0.1} & \multicolumn{1}{r}{0.0} & \multicolumn{1}{r}{0.1} & \multicolumn{1}{r}{0.1} \\ \hline
\multicolumn{1}{l|}{\multirow{4}{*}{\textbf{\textbf{Meemi}\textsubscript{VM}}}} & \multicolumn{1}{l|}{\textbf{8K}} & \multicolumn{1}{r}{\textbf{9.8}} & \multicolumn{1}{r}{\textbf{21.3}} & \multicolumn{1}{r|}{\textbf{26.9}} & \multicolumn{1}{r}{\textbf{10.6}} & \multicolumn{1}{r}{\textbf{20.0}} & \multicolumn{1}{r|}{\textbf{25.6}} & \multicolumn{1}{r}{\textbf{3.7}} & \multicolumn{1}{r}{\textbf{9.6}} & \multicolumn{1}{r|}{\textbf{13.2}} & \multicolumn{1}{r}{1.3} & \multicolumn{1}{r}{3.6} & \multicolumn{1}{r|}{5.5} & \multicolumn{1}{r}{0.0} & \multicolumn{1}{r}{0.1} & \multicolumn{1}{r}{0.1} \\
\multicolumn{1}{l|}{} & \multicolumn{1}{l|}{\textbf{1K}} & \multicolumn{1}{r}{8.3} & \multicolumn{1}{r}{17.7} & \multicolumn{1}{r|}{22.6} & \multicolumn{1}{r}{8.6} & \multicolumn{1}{r}{18.2} & \multicolumn{1}{r|}{23.6} & \multicolumn{1}{r}{3.0} & \multicolumn{1}{r}{7.5} & \multicolumn{1}{r|}{10.6} & \multicolumn{1}{r}{0.5} & \multicolumn{1}{r}{2.4} & \multicolumn{1}{r|}{3.7} & \multicolumn{1}{r}{0.0} & \multicolumn{1}{r}{0.0} & \multicolumn{1}{r}{0} \\
\multicolumn{1}{l|}{} & \multicolumn{1}{l|}{\textbf{100}} & \multicolumn{1}{r}{0.0} & \multicolumn{1}{r}{0.0} & \multicolumn{1}{r|}{0.0} & \multicolumn{1}{r}{0.0} & \multicolumn{1}{r}{0.0} & \multicolumn{1}{r|}{0.0} & \multicolumn{1}{r}{0.0} & \multicolumn{1}{r}{0.0} & \multicolumn{1}{r|}{0.0} & \multicolumn{1}{r}{0.0} & \multicolumn{1}{r}{0.0} & \multicolumn{1}{r|}{0.0} & \multicolumn{1}{r}{0.0} & \multicolumn{1}{r}{0.0} & \multicolumn{1}{r}{0.0} \\
\multicolumn{1}{l|}{} & \multicolumn{1}{l|}{\textbf{Identical}} & \multicolumn{1}{r}{3.8} & \multicolumn{1}{r}{9.1} & \multicolumn{1}{r|}{11.8} & \multicolumn{1}{r}{6.6} & \multicolumn{1}{r}{14.2} & \multicolumn{1}{r|}{18.2} & \multicolumn{1}{r}{2.0} & \multicolumn{1}{r}{4.1} & \multicolumn{1}{r|}{5.9} & \multicolumn{1}{r}{0.0} & \multicolumn{1}{r}{0.1} & \multicolumn{1}{r|}{0.2} & \multicolumn{1}{r}{0.1} & \multicolumn{1}{r}{0.3} & \multicolumn{1}{r}{0.5} \\ \hline
\multicolumn{1}{l|}{\multirow{4}{*}{\textbf{\textbf{Meemi}\textsubscript{MS}}}} & \multicolumn{1}{l|}{\textbf{8K}} & \multicolumn{1}{r}{9.5} & \multicolumn{1}{r}{20.5} & \multicolumn{1}{r|}{26.3} & \multicolumn{1}{r}{9.5} & \multicolumn{1}{r}{19.1} & \multicolumn{1}{r|}{24.5} & \multicolumn{1}{r}{3.0} & \multicolumn{1}{r}{7.6} & \multicolumn{1}{r|}{11.1} & \multicolumn{1}{r}{\textbf{1.5}} & \multicolumn{1}{r}{\textbf{4.3}} & \multicolumn{1}{r|}{\textbf{6.4}} & \multicolumn{1}{r}{0.0} & \multicolumn{1}{r}{0.1} & \multicolumn{1}{r}{0.2} \\
\multicolumn{1}{l|}{} & \multicolumn{1}{l|}{\textbf{1K}} & \multicolumn{1}{r}{7.6} & \multicolumn{1}{r}{16.9} & \multicolumn{1}{r|}{22.3} & \multicolumn{1}{r}{7.8} & \multicolumn{1}{r}{15.9} & \multicolumn{1}{r|}{21} & \multicolumn{1}{r}{1.7} & \multicolumn{1}{r}{4.1} & \multicolumn{1}{r|}{6.2} & \multicolumn{1}{r}{0.8} & \multicolumn{1}{r}{2.3} & \multicolumn{1}{r|}{3.7} & \multicolumn{1}{r}{0} & \multicolumn{1}{r}{0.0} & \multicolumn{1}{r}{0.0} \\
\multicolumn{1}{l|}{} & \multicolumn{1}{l|}{\textbf{100}} & \multicolumn{1}{r}{0.0} & \multicolumn{1}{r}{0.0} & \multicolumn{1}{r|}{0.0} & \multicolumn{1}{r}{0.0} & \multicolumn{1}{r}{0.0} & \multicolumn{1}{r|}{0.0} & \multicolumn{1}{r}{0.0} & \multicolumn{1}{r}{0.0} & \multicolumn{1}{r|}{0.0} & \multicolumn{1}{r}{0.0} & \multicolumn{1}{r}{0.0} & \multicolumn{1}{r|}{0.0} & \multicolumn{1}{r}{0.0} & \multicolumn{1}{r}{0.0} & \multicolumn{1}{r}{0.0} \\
\multicolumn{1}{l|}{} & \multicolumn{1}{l|}{\textbf{Identical}} & \multicolumn{1}{r}{2.9} & \multicolumn{1}{r}{5.3} & \multicolumn{1}{r|}{6.7} & \multicolumn{1}{r}{3.5} & \multicolumn{1}{r}{9.9} & \multicolumn{1}{r|}{13.2} & \multicolumn{1}{r}{1.2} & \multicolumn{1}{r}{2.8} & \multicolumn{1}{r|}{3.6} & \multicolumn{1}{r}{0.2} & \multicolumn{1}{r}{0.3} & \multicolumn{1}{r|}{0.3} & \multicolumn{1}{r}{0.0} & \multicolumn{1}{r}{0.1} & \multicolumn{1}{r}{0.2} \\ \hline
 
\end{tabular}
}
\caption{Bilingual dictionary induction results in the test sets of \protect\newcite{conneau2018word}.}
\label{tab:dictionary}
\end{table*}

\begin{table*}[!t]
\renewcommand{\arraystretch}{0.96}
\resizebox{\textwidth}{!}{
\begin{tabular}{clrrrrrrrr}
\hline

\multicolumn{10}{c}{\texttt{\textbf{Wikipedia}}} \\ \hline \hline
\multicolumn{1}{c|}{\multirow{2}{*}{\textbf{Model}}} & \multicolumn{1}{c|}{\multirow{2}{*}{\textbf{Dictionary}}} & \multicolumn{2}{c}{\textbf{English-Spanish}} & \multicolumn{2}{c|}{\textbf{English-Italian}} & \multicolumn{2}{c|}{\textbf{English-German}} & \multicolumn{2}{c|}{\textbf{English-Farsi}} \\ \cline{3-10} 
\multicolumn{1}{c|}{} & \multicolumn{1}{c|}{} & \multicolumn{1}{c}{\textbf{Pearson}} & \multicolumn{1}{c|}{\textbf{Spearman}} & \multicolumn{1}{c}{\textbf{Pearson}} & \multicolumn{1}{c|}{\textbf{Spearman}} & \multicolumn{1}{c}{\textbf{Pearson}} & \multicolumn{1}{c|}{\textbf{Spearman}} & \multicolumn{1}{c}{\textbf{Pearson}} & \multicolumn{1}{c}{\textbf{Spearman}} \\ \hline
\multicolumn{1}{c|}{\multirow{5}{*}{\textbf{\texttt{VecMap}}}} & \multicolumn{1}{l|}{\textbf{8K}} & 72.1 & \multicolumn{1}{r|}{71.8} & 71.2 & \multicolumn{1}{r|}{70.6} & 70.0 & \multicolumn{1}{r|}{69.3} & 63.7 & 61.7 \\
\multicolumn{1}{c|}{} & \multicolumn{1}{l|}{\textbf{1K}} & 72.1 & \multicolumn{1}{r|}{71.8} & 71.2 & \multicolumn{1}{r|}{70.6} & 70.0 & \multicolumn{1}{r|}{69.3} & 63.9 & 61.9 \\
\multicolumn{1}{c|}{} & \multicolumn{1}{l|}{\textbf{100}} & 72.1 & \multicolumn{1}{r|}{71.8} & 71.2 & \multicolumn{1}{r|}{70.6} & 70.0 & \multicolumn{1}{r|}{69.3} & 63.9 & 62.0 \\
\multicolumn{1}{c|}{} & \multicolumn{1}{l|}{\textbf{Identical}} & 72.1 & \multicolumn{1}{r|}{71.8} & 71.2 & \multicolumn{1}{r|}{70.6} & 70.0 & \multicolumn{1}{r|}{69.3} & 63.8 & 61.9 \\
\multicolumn{1}{c|}{} & \multicolumn{1}{l|}{\textbf{Unsupervised}} & 72.1 & \multicolumn{1}{r|}{71.8} & 71.2 & \multicolumn{1}{r|}{70.6} & 70.0 & \multicolumn{1}{r|}{69.3} & 63.4 & 61.3 \\ \hline
\multicolumn{1}{c|}{\multirow{5}{*}{\textbf{\texttt{MUSE}}}} & \multicolumn{1}{l|}{\textbf{8K}} & 72.0 & \multicolumn{1}{r|}{72.6} & 70.7 & \multicolumn{1}{r|}{70.9} & 68.8 & \multicolumn{1}{r|}{68.9} & 59.2 & 58.7 \\
\multicolumn{1}{c|}{} & \multicolumn{1}{l|}{\textbf{1K}} & 71.9 & \multicolumn{1}{r|}{72.4} & 70.6 & \multicolumn{1}{r|}{70.7} & 68.6 & \multicolumn{1}{r|}{68.7} & 58.7 & 58.4 \\
\multicolumn{1}{c|}{} & \multicolumn{1}{l|}{\textbf{100}} & 65.1 & \multicolumn{1}{r|}{66.3} & 63.0 & \multicolumn{1}{r|}{63.6} & 44.7 & \multicolumn{1}{r|}{49.9} & 47.7 & 52.1 \\
\multicolumn{1}{c|}{} & \multicolumn{1}{l|}{\textbf{Identical}} & 71.0 & \multicolumn{1}{r|}{71.9} & 69.9 & \multicolumn{1}{r|}{70.5} & 68.1 & \multicolumn{1}{r|}{68.4} & 47.9 & 51.3 \\
\multicolumn{1}{c|}{} & \multicolumn{1}{l|}{\textbf{Unsupervised}} & 72.2 & \multicolumn{1}{r|}{72.6} & 71.0 & \multicolumn{1}{r|}{71.2} & 68.7 & \multicolumn{1}{r|}{68.9} & 8.0 & 6.5 \\ \hline
\multicolumn{1}{c|}{\multirow{4}{*}{\textbf{\begin{tabular}[c]{@{}c@{}}\textbf{Meemi}\textsubscript{VM}\end{tabular}}}} & \multicolumn{1}{l|}{\textbf{8K}} & 72.5 & \multicolumn{1}{r|}{71.9} & 71.8 & \multicolumn{1}{r|}{70.9} & \textbf{70.9} & \multicolumn{1}{r|}{\textbf{70.3}} & \textbf{65.1} & \textbf{63.4} \\
\multicolumn{1}{c|}{} & \multicolumn{1}{l|}{\textbf{1K}} & 70.1 & \multicolumn{1}{r|}{69.6} & 69.7 & \multicolumn{1}{r|}{69.0} & 67.6 & \multicolumn{1}{r|}{66.8} & 5.5 & 5.8 \\
\multicolumn{1}{c|}{} & \multicolumn{1}{l|}{\textbf{100}} & 0.0 & \multicolumn{1}{r|}{0.0} & 5.1 & \multicolumn{1}{r|}{5.0} & 4.1 & \multicolumn{1}{r|}{3.3} & 6.8 & 6.5 \\
\multicolumn{1}{c|}{} & \multicolumn{1}{l|}{\textbf{Identical}} & 71.0 & \multicolumn{1}{r|}{70.4} & 69.4 & \multicolumn{1}{r|}{68.7} & 69.3 & \multicolumn{1}{r|}{68.6} & 56.1 & 54.1 \\ \hline
\multicolumn{1}{c|}{\multirow{4}{*}{\textbf{\begin{tabular}[c]{@{}c@{}}\textbf{Meemi}\textsubscript{MS}\end{tabular}}}} & \multicolumn{1}{l|}{\textbf{8K}} & \textbf{73.1} & \multicolumn{1}{r|}{\textbf{72.9}} & \textbf{72.4} & \multicolumn{1}{r|}{\textbf{71.9}} & 70.7 & \multicolumn{1}{r|}{70.1} & 64.1 & 62.0 \\
\multicolumn{1}{c|}{} & \multicolumn{1}{l|}{\textbf{1K}} & 70.4 & \multicolumn{1}{r|}{70.3} & 69.6 & \multicolumn{1}{r|}{69.6} & 66.7 & \multicolumn{1}{r|}{66.5} & 5.0 & 4.2 \\
\multicolumn{1}{c|}{} & \multicolumn{1}{l|}{\textbf{100}} & 2.7 & \multicolumn{1}{r|}{1.4} & 0.0 & \multicolumn{1}{r|}{0.2} & 6.3 & \multicolumn{1}{r|}{6.0} & 0.0 & 0.0 \\
\multicolumn{1}{c|}{} & \multicolumn{1}{l|}{\textbf{Identical}} & 70.6 & \multicolumn{1}{r|}{70.9} & 68.7 & \multicolumn{1}{r|}{68.8} & 68.0 & \multicolumn{1}{r|}{67.7} & 46.6 & 48.2 \\ \hline
\hline

\multicolumn{10}{c}{\texttt{\textbf{Web corpora}}} \\ \hline \hline
\multicolumn{1}{c|}{\multirow{2}{*}{\textbf{Model}}} & \multicolumn{1}{l|}{\multirow{2}{*}{\textbf{Dictionary}}} & \multicolumn{2}{c|}{\textbf{English-Spanish}} & \multicolumn{2}{c|}{\textbf{English-Italian}} & \multicolumn{2}{c|}{\textbf{English-German}} & \multicolumn{2}{c}{\textbf{English-Farsi}} \\ \cline{3-10} 
\multicolumn{1}{c|}{} & \multicolumn{1}{l|}{} & \multicolumn{1}{l}{\textbf{Pearson}} & \multicolumn{1}{l|}{\textbf{Spearman}} & \multicolumn{1}{l}{\textbf{Pearson}} & \multicolumn{1}{l|}{\textbf{Spearman}} & \multicolumn{1}{l}{\textbf{Pearson}} & \multicolumn{1}{l|}{\textbf{Spearman}} & \multicolumn{1}{l}{\textbf{Pearson}} & \multicolumn{1}{l}{\textbf{Spearman}} \\ \hline
\multicolumn{1}{c|}{\multirow{5}{*}{\textbf{\texttt{VecMap}}}} & \multicolumn{1}{l|}{\textbf{8K}} & 71.0 & \multicolumn{1}{r|}{70.6} & 69.2 & \multicolumn{1}{r|}{68.8} & 70.9 & \multicolumn{1}{r|}{70.4} & 35.9 & 33.5 \\
\multicolumn{1}{c|}{} & \multicolumn{1}{l|}{\textbf{1K}} & 71.0 & \multicolumn{1}{r|}{70.6} & 69.3 & \multicolumn{1}{r|}{68.8} & 70.9 & \multicolumn{1}{r|}{70.4} & 35.9 & 33.5 \\
\multicolumn{1}{c|}{} & \multicolumn{1}{l|}{\textbf{100}} & 71.0 & \multicolumn{1}{r|}{70.4} & 69.2 & \multicolumn{1}{r|}{68.8} & 71.0 & \multicolumn{1}{r|}{70.5} & 35.9 & 33.5 \\
\multicolumn{1}{c|}{} & \multicolumn{1}{l|}{\textbf{Identical}} & 71.0 & \multicolumn{1}{r|}{70.6} & 69.3 & \multicolumn{1}{r|}{68.8} & 70.9 & \multicolumn{1}{r|}{70.4} & 35.9 & 33.0 \\
\multicolumn{1}{c|}{} & \multicolumn{1}{l|}{\textbf{Unsupervised}} & 71.1 & \multicolumn{1}{r|}{70.5} & 69.2 & \multicolumn{1}{r|}{68.8} & 70.9 & \multicolumn{1}{r|}{70.4} & 35.7 & 33.4 \\ \hline
\multicolumn{1}{c|}{\multirow{5}{*}{\textbf{\texttt{MUSE}}}} & \multicolumn{1}{l|}{\textbf{8K}} & 71.9 & \multicolumn{1}{r|}{71.9} & 70.4 & \multicolumn{1}{r|}{70.4} & 70.5 & \multicolumn{1}{r|}{70.2} & 29.7 & 23.9 \\
\multicolumn{1}{c|}{} & \multicolumn{1}{l|}{\textbf{1K}} & 71.6 & \multicolumn{1}{r|}{71.5} & 69.5 & \multicolumn{1}{r|}{69.4} & 70.3 & \multicolumn{1}{r|}{70.0} & 28.3 & 22.3 \\
\multicolumn{1}{c|}{} & \multicolumn{1}{l|}{\textbf{100}} & 71.7 & \multicolumn{1}{r|}{71.6} & 67.4 & \multicolumn{1}{r|}{67.4} & 68.5 & \multicolumn{1}{r|}{68.8} & 6.3 & 7.4 \\
\multicolumn{1}{c|}{} & \multicolumn{1}{l|}{\textbf{Identical}} & 69.9 & \multicolumn{1}{r|}{70.1} & 67.3 & \multicolumn{1}{r|}{67.5} & 70.1 & \multicolumn{1}{r|}{69.7} & 17.5 & 14.5 \\
\multicolumn{1}{c|}{} & \multicolumn{1}{l|}{\textbf{Unsupervised}} & 71.7 &\multicolumn{1}{r|}{71.6} & 69.4	 & \multicolumn{1}{r|}{69.4}   & 70.3 & \multicolumn{1}{r|}{70.0} & 29.6 & 23.8 \\ \hline
\multicolumn{1}{c|}{\multirow{4}{*}{\textbf{\begin{tabular}[c]{@{}c@{}}\textbf{Meemi}\textsubscript{VM}\end{tabular}}}} & \multicolumn{1}{l|}{\textbf{8K}} & 71.5 & \multicolumn{1}{r|}{70.9} & 70.4 & \multicolumn{1}{r|}{70.0} & 72.3 & \multicolumn{1}{r|}{71.8} & \textbf{40.2} & \textbf{39.0} \\
\multicolumn{1}{c|}{} & \multicolumn{1}{l|}{\textbf{1K}} & 69.1 & \multicolumn{1}{r|}{68.6} & 68.2 & \multicolumn{1}{r|}{67.8} & 70.9 & \multicolumn{1}{r|}{70.2} & 1.2 & 0.7 \\
\multicolumn{1}{c|}{} & \multicolumn{1}{l|}{\textbf{100}} & 0.0 & \multicolumn{1}{r|}{0.0} & 0.0 & \multicolumn{1}{r|}{0.0} & 0.0 & \multicolumn{1}{r|}{0.0} & 3.2 & 3.0 \\
\multicolumn{1}{c|}{} & \multicolumn{1}{l|}{\textbf{Identical}} & 70.1 & \multicolumn{1}{r|}{69.5} & 69.2 & \multicolumn{1}{r|}{68.4} & 71.2 & \multicolumn{1}{r|}{70.6} & 31.6 & 28.5 \\ \hline
\multicolumn{1}{c|}{\multirow{4}{*}{\textbf{\begin{tabular}[c]{@{}c@{}}\textbf{Meemi}\textsubscript{MS}\end{tabular}}}} & \multicolumn{1}{l|}{\textbf{8K}} & \textbf{72.5} & \multicolumn{1}{r|}{\textbf{72.3}} & \textbf{71.5} & \multicolumn{1}{r|}{\textbf{71.1}} & \textbf{72.5} & \multicolumn{1}{r|}{\textbf{72.1}} & 36.4 & 33.0 \\
\multicolumn{1}{c|}{} & \multicolumn{1}{l|}{\textbf{1K}} & 70.0 & \multicolumn{1}{r|}{69.7} & 68.9 & \multicolumn{1}{r|}{68.6} & 70.3 & \multicolumn{1}{r|}{69.9} & 0.0 & 0.0 \\
\multicolumn{1}{c|}{} & \multicolumn{1}{l|}{\textbf{100}} & 1.7 & \multicolumn{1}{r|}{0.1} & 1.6 & \multicolumn{1}{r|}{2.3} & 0.0 & \multicolumn{1}{r|}{0.0} & 0.3 & 0.0 \\
\multicolumn{1}{c|}{} & \multicolumn{1}{l|}{\textbf{Identical}} & 69.2 & \multicolumn{1}{r|}{69.1} & 67.4 & \multicolumn{1}{r|}{66.9} & 70.1 & \multicolumn{1}{r|}{69.4} & 17.3 & 14.5 \\ \hline \hline

\multicolumn{10}{c}{\texttt{\textbf{Social media}}} \\ \hline \hline
\multicolumn{1}{c|}{\multirow{2}{*}{\textbf{Model}}} & \multicolumn{1}{l|}{\multirow{2}{*}{\textbf{Dictionary}}} & \multicolumn{2}{c|}{\textbf{English-Spanish}} & \multicolumn{2}{c|}{\textbf{English-Italian}} & \multicolumn{2}{c|}{\textbf{English-German}} & \multicolumn{2}{c}{\textbf{English-Farsi}} \\ \cline{3-10} 
\multicolumn{1}{c|}{} & \multicolumn{1}{l|}{} & \multicolumn{1}{l}{\textbf{Pearson}} & \multicolumn{1}{l|}{\textbf{Spearman}} & \multicolumn{1}{l}{\textbf{Pearson}} & \multicolumn{1}{l|}{\textbf{Spearman}} & \multicolumn{1}{l}{\textbf{Pearson}} & \multicolumn{1}{l|}{\textbf{Spearman}} & \multicolumn{1}{l}{\textbf{Pearson}} & \multicolumn{1}{l}{\textbf{Spearman}} \\ \hline
\multicolumn{1}{c|}{\multirow{5}{*}{\textbf{\texttt{VecMap}}}} & \multicolumn{1}{l|}{\textbf{8K}} & 48.5 & \multicolumn{1}{r|}{47.4} & 53.9 & \multicolumn{1}{r|}{51.8} & 51.3 & \multicolumn{1}{r|}{49.5} & 31.1 & 30.3 \\
\multicolumn{1}{c|}{} & \multicolumn{1}{l|}{\textbf{1K}} & 48.7 & \multicolumn{1}{r|}{47.7} & 53.7 & \multicolumn{1}{r|}{51.6} & 51.7 & \multicolumn{1}{r|}{50.2} & 30.3 & 29.6 \\
\multicolumn{1}{c|}{} & \multicolumn{1}{l|}{\textbf{100}} & 49.8 & \multicolumn{1}{r|}{48.9} & 54.0 & \multicolumn{1}{r|}{51.7} & 51.0 & \multicolumn{1}{r|}{49.5} & 25.7 & 25.4 \\
\multicolumn{1}{c|}{} & \multicolumn{1}{l|}{\textbf{Identical}} & 48.4 & \multicolumn{1}{r|}{47.1} & 54.2 & \multicolumn{1}{r|}{51.9} & 51.8 & \multicolumn{1}{r|}{50.3} & 27.8 & 26.5 \\
\multicolumn{1}{c|}{} & \multicolumn{1}{l|}{\textbf{Unsupervised}} & 48.0 & \multicolumn{1}{r|}{46.9} & 53.8 & \multicolumn{1}{r|}{51.5} & 30.1 & \multicolumn{1}{r|}{31.2} & 4.2 & 2.4 \\ \hline
\multicolumn{1}{c|}{\multirow{5}{*}{\textbf{\texttt{MUSE}}}} & \multicolumn{1}{l|}{\textbf{8K}} & 48.8 & \multicolumn{1}{r|}{47.6} & 51.0 & \multicolumn{1}{r|}{49.3} & 48.5 & \multicolumn{1}{r|}{48.6} & 43.3 & 42.2 \\
\multicolumn{1}{c|}{} & \multicolumn{1}{l|}{\textbf{1K}} & 46.6 & \multicolumn{1}{r|}{45.5} & 49.7 & \multicolumn{1}{r|}{47.8} & 44.8 & \multicolumn{1}{r|}{45.7} & 38.7 & 38.9 \\
\multicolumn{1}{c|}{} & \multicolumn{1}{l|}{\textbf{100}} & 35.8 & \multicolumn{1}{r|}{36.9} & 29.6 & \multicolumn{1}{r|}{31.3} & 30.7 & \multicolumn{1}{r|}{34.0} & 20.8 & 21.3 \\
\multicolumn{1}{c|}{} & \multicolumn{1}{l|}{\textbf{Identical}} & 48.1 & \multicolumn{1}{r|}{47.7} & 50.1 & \multicolumn{1}{r|}{49.8} & 45.6 & \multicolumn{1}{r|}{46.8} & 30.5 & 32.4 \\
\multicolumn{1}{c|}{} & \multicolumn{1}{l|}{\textbf{Unsupervised}} & 9.9 & \multicolumn{1}{r|}{10.9} & 50.7 & \multicolumn{1}{r|}{49.7} & 12.4 & \multicolumn{1}{r|}{13.0} & 6.9 & 4.7 \\ \hline
\multicolumn{1}{c|}{\multirow{4}{*}{\textbf{\begin{tabular}[c]{@{}c@{}}\textbf{Meemi}\textsubscript{VM}\end{tabular}}}} & \multicolumn{1}{l|}{\textbf{8K}} & 51.2 & \multicolumn{1}{r|}{50.1} & \textbf{56.1} & \multicolumn{1}{r|}{\textbf{53.6}} & \textbf{55.0} & \multicolumn{1}{r|}{\textbf{53.8}} & 45.2 & 43.1  \\
\multicolumn{1}{c|}{} & \multicolumn{1}{l|}{\textbf{1K}} & 49.7 & \multicolumn{1}{r|}{48.6} & 55.4 & \multicolumn{1}{r|}{52.8} & 52.8 & \multicolumn{1}{r|}{51.3} & 3.8 & 3.8 \\
\multicolumn{1}{c|}{} & \multicolumn{1}{l|}{\textbf{100}} & 2.4 & \multicolumn{1}{r|}{2.0} & 2.1 & \multicolumn{1}{r|}{2.8} & 0.0 & \multicolumn{1}{r|}{0.0} & 5.6 & 5.1 \\
\multicolumn{1}{c|}{} & \multicolumn{1}{l|}{\textbf{Identical}} & 51.7 & \multicolumn{1}{r|}{50.1} & 56.2 & \multicolumn{1}{r|}{53.4} & 52.7 & \multicolumn{1}{r|}{51.3} & 29.9 & 28.6 \\ \hline
\multicolumn{1}{c|}{\multirow{4}{*}{\textbf{\begin{tabular}[c]{@{}c@{}}\textbf{Meemi}\textsubscript{MS}\end{tabular}}}} & \multicolumn{1}{l|}{\textbf{8K}} & \textbf{51.8} & \multicolumn{1}{r|}{\textbf{50.4}} & 54.8 & \multicolumn{1}{r|}{52.5} & 53.1 & \multicolumn{1}{r|}{52.0} & \textbf{48.9} & \textbf{46.6} \\
\multicolumn{1}{c|}{} & \multicolumn{1}{l|}{\textbf{1K}} & 49.5 & \multicolumn{1}{r|}{47.9} & 53.1 & \multicolumn{1}{r|}{50.7} & 48.4 & \multicolumn{1}{r|}{47.2} & 0.0 & 0.0 \\
\multicolumn{1}{c|}{} & \multicolumn{1}{l|}{\textbf{100}} & 5.2 & \multicolumn{1}{r|}{5.4} & 0.0 & \multicolumn{1}{r|}{0.0} & 6.2 & \multicolumn{1}{r|}{6.8} & 0.0 & 0.0 \\
\multicolumn{1}{c|}{} & \multicolumn{1}{l|}{\textbf{Identical}} & 49.6 & \multicolumn{1}{r|}{48.2} & 53.1 & \multicolumn{1}{r|}{51.4} & 48.4 & \multicolumn{1}{r|}{47.7} & 30.5 & 30.6

\end{tabular}
}
\caption{Cross-lingual word similarity results in the SemEval-17 dataset \protect\cite{semeval2017similarity}.}
\label{tab:sim}
\end{table*}

\section{Bibliographical References}
\bibliographystyle{lrec}
\bibliography{biblio}

\end{document}